\documentclass[fleqn,10pt,twocolumn]{wlscirep}
\newcommand{\dbar}{d\hspace*{-0.08em}\bar{}\vspace*{-0.1em}}
\usepackage{lipsum,multicol}

\usepackage[utf8]{inputenc}
\usepackage[T1]{fontenc}
\usepackage{graphicx,float}
\usepackage{subfig}
\usepackage{cite}
\usepackage{multirow}
\usepackage{amsmath,amssymb,amsfonts}
\usepackage{algorithmic}
\usepackage{lmodern}
\usepackage{titlesec}
\usepackage{times}
\usepackage[symbol]{footmisc}
\usepackage{hyperref}
\usepackage{lettrine}
\usepackage{xcolor}
\usepackage{color,soul}
\usepackage[utf8]{inputenc}
\usepackage{authblk}
\usepackage{hyperref}
\newbox{\myorcidaffilbox}
\sbox{\myorcidaffilbox}{\large\includegraphics[height=4mm]{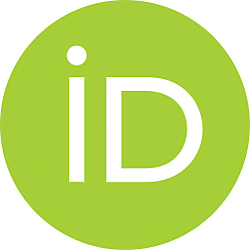}}
\newcommand{\orcidaffil}[1]{\href{https://orcid.org/#1}{\usebox{\myorcidaffilbox}}}
 
\newbox{\mybox}
\sbox{\mybox}{\large\includegraphics[height=2.5mm]{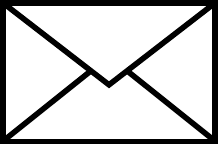}}
\newcommand{\boxaffil}[1]{\href{https://orcid.org/#1}{\usebox{\mybox}}}
\usepackage{lipsum,microtype}
\usepackage{subfig}
\usepackage{framed,multirow}
\usepackage[symbol]{footmisc}
\usepackage{amssymb}
\usepackage{latexsym}
\usepackage{url}
\usepackage{xcolor}
\usepackage{hyperref}
\usepackage{geometry}
\usepackage{fleqn}
\usepackage{newtxtext,newtxmath}
\usepackage{amsfonts}
\usepackage{scalerel}
\usepackage[edges]{forest}
\forestset{declare boolean={switch me}{0}}
\usepackage{graphicx,float}
\usepackage[export]{adjustbox}
\usepackage{graphicx}
\usepackage{wrapfig,booktabs}
\usepackage{multirow,  makecell, booktabs, tabularx}
\usepackage{tabularx,ragged2e}
\usepackage[export]{adjustbox}
\usepackage{amsmath}

\usepackage{hyperref}
\hypersetup{
    colorlinks=true,
    linkcolor=red,
    filecolor=magenta, 
    urlcolor=blue,
}
\usepackage{blindtext}
\usepackage{etoolbox}
\usepackage{pgfplots}
\usepackage{pgfplotstable}

\title{Traffic-Net: 3D Traffic Monitoring Using a Single Camera}
\author{Mahdi Rezaei}
\date{February 2021}

\date{}
\author{
Mahdi Rezaei \orcidaffil{0000-0003-3892-421X}  $^{1, \star, \, \boxaffil{}}$ , \, Mohsen Azarmi \orcidaffil{0000-0003-0737-9204}  $^{2, \star}$, \, Farzam Mohammad Pour Mir \orcidaffil{0000-0002-5377-8380}  $^{3, \star}$ \\ 
$^1$ Institute for Transport Studies, The University of Leeds, Leeds, LS2 9JT, UK\\ 
$^2$ Department of Computer Engineering, Qazvin University, Qazvin, IR\\
$^3$ Faculty of Computer Engineering, Tehran Azad University, Science \& Research Branch, IR\\ 
$^1$ \href{mailto:m.rezaei@leeds.ac.uk}{m.rezaei@leeds.ac.uk}  \, $^2$ \href{mailto:m.azarmi@qiau.ac.ir}{m.azarmi@qiau.ac.ir} \, $^3$ \href{mailto:f.mohammadpour@srbiau.ac.ir}{f.mohammadpour@srbiau.ac.ir}  
}

\vspace{-5mm}
\begin{abstract}
Computer Vision has played a major role in Intelligent Transportation Systems (ITS) and traffic surveillance. Along with the rapidly growing automated vehicles and crowded cities, the automated and advanced traffic management systems (ATMS) using video surveillance infrastructures have been evolved by the implementation of Deep Neural Networks. 
In this research, we provide a practical platform for real-time traffic monitoring, including 3D vehicle/pedestrian detection, speed detection, trajectory estimation, congestion detection, as well as monitoring the interaction of vehicles and pedestrians, all using a single CCTV traffic camera.
We adapt a custom YOLOv5 deep neural network model for vehicle/pedestrian detection and an enhanced SORT tracking algorithm. 
For the first time, a hybrid satellite-ground based inverse perspective mapping (SG-IPM) method for camera auto-calibration is also developed which leads to an accurate 3D object detection and visualisation.
We also develop a hierarchical traffic modelling solution based on short- and long-term temporal video data stream to understand the traffic flow, bottlenecks, and risky spots for vulnerable road users. 
Several experiments on real-world scenarios and comparisons with state-of-the-art are conducted using various traffic monitoring datasets, including MIO-TCD, UA-DETRAC and GRAM-RTM collected from highways, intersections, and urban areas under different lighting and weather conditions.\\

\vspace{-2mm}

\end{abstract}

\pgfplotsset{compat=1.17}
\begin{document}
%\maketitle
\flushbottom
\maketitle
\thispagestyle{empty}

\footnote[0]{$^\star$ The authors contributed equally to this study.\\
$^{\href{https://orcid.org/0000-0003-3892-421X}{\includegraphics[width=3mm]{drawing.pdf}}}$ Corresponding Author: \href{mailto:m.rezaei@leeds.ac.uk}{m.rezaei@leeds.ac.uk} (M. Rezaei) }
\vspace{-12mm}

%=========================================
\section{Introduction}
\lettrine[lines=3]{\textcolor[rgb]{0.4,0.4,0.4}S}{\, mart} video surveillance systems are becoming a common technology for traffic monitoring and congestion management. Parallel to the technology improvements, the complexity of traffic scenes for automated traffic surveillance has also increased due to multiple factors such as urban developments, the mixture of classic and autonomous vehicles, population growth, and the increasing number of pedestrians and road users \cite{nam2018}.  The rapidly growing number of surveillance cameras (over 20 million CCD cameras only in USA and UK) in the arteries of cities, crowded places, roads, intersections, and highways, demonstrator the importance of video surveillance for city councils, authorities and governments \cite{sheng2021surveilling}.
A large network of interconnected surveillance cameras can provide a special platform for further studies on traffic management and urban planning \cite{olatunji2019video}. However, in such a dense and complex road environments, the conventional monitoring of road condition is a very tedious, time-consuming, yet less accurate approach than automated computer vision and AI-based solutions. Hence, automated video surveillance has been researched for many years to gradually replace the humans with computers that can analyse the live traffic and provide effective solutions to maintain transportation safety and sustainability \cite{Mondal2020}.
\begin{figure}[t!]
\vspace{-3mm}
\centering
%\subfloat[3D object detection and speed estimation]
{\resizebox{7.2cm}{!}{\includegraphics{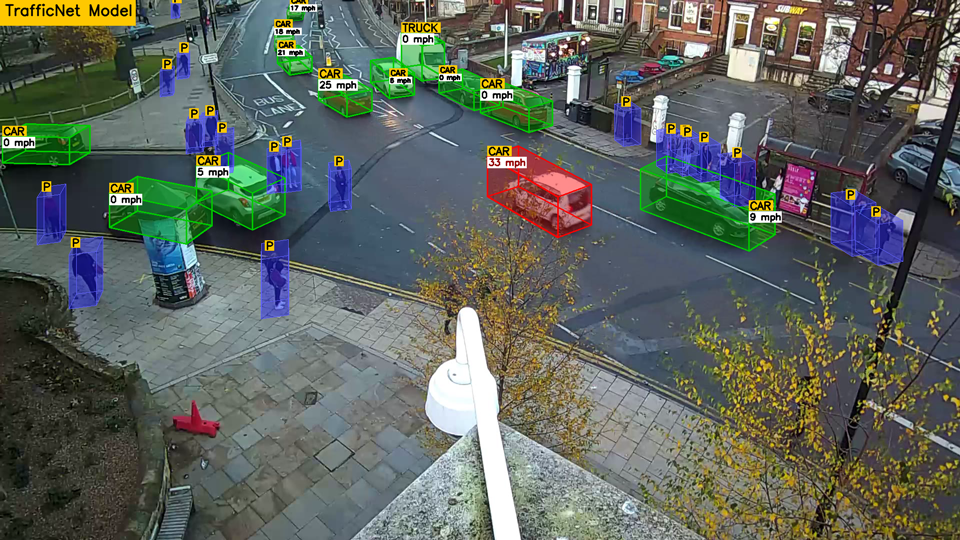}\label{tiser_3d}}}\vspace{1pt}
%\subfloat[Environment modelling]
{\resizebox{6.1cm}{!}{\includegraphics{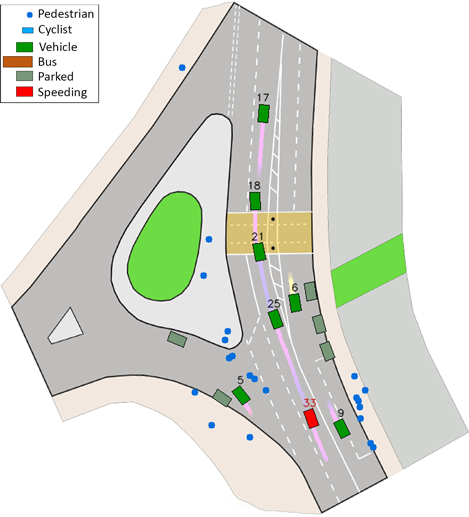}\label{tiser_env}}}
\vspace{-2.9mm}
\caption{Top: 3D object detection and speed estimation. Bottom: Digital twin and modelling of the same scene.}
\label{fig-tiser}
\vspace{-4mm}
\end{figure}

Computer Vision is one of the most investigated technologies for automated video surveillance inspired by the human visual mechanism.
The technology aims to enable computers to analyse and interpret the content of digital videos. 
In automated traffic monitoring systems (ATMS) and Intelligent Transportation Systems (ITS), computer vision can extract a wide range of information from the traffic scenes \cite{poddar2016automated}. 

Vehicle type recognition, vehicle counting, speed estimation, tracking, and trajectory estimation are examples of automated traffic scene analysis.
Figure \ref{fig-tiser} represents a sample scenario of road complexities including interactions between road users (pedestrians, vehicles, cyclists), moving trajectories, speed detection, and the density of the road users in various points of the road scene. 
Figure \ref{fig-tiser}, top row shows a 3D road-user detection and localisation, and the bottom row shows the bird's eye view mapping and digital twin modelling of the same scene, after camera calibration and inverse perspective mapping (IPM). 

In such highly dynamic environments, the ability of real-time processing and accurate detection of multiple events is crucial \cite{Hu2004}.
Furthermore, an efficient traffic monitoring system should be capable of working with a grid of various interconnected cameras on different urban locations, where each camera may have a different resolution, viewing angles, height, or focal length.
This requires calibration of each and every single camera based on the intrinsic camera parameters and the mounting spec of each camera. 

Although various methods of camera calibration such as vanishing-based techniques \cite{vanishing2018} and multi-point calibrations techniques \cite{IPM2015} have been introduced for bird's eye view mapping, fewer investigations have been conducted in the community to introduce automated calibration methods.

The heart of an ATMS is the vehicle and pedestrian identification, and in the field of computer vision, this task is handled by object detection algorithms and tracking techniques \cite{brunetti2018computer}.

In the past decade, deployment of Deep Neural Networks (DNN) has led to significant advances in indoor object detection. 
The effectiveness and the accuracy of these contemporary improvements should be investigated for the particular application of traffic monitoring in a complex, dynamic, noisy and crowded environment. 

Further challenges such as bad weather conditions, challenging lighting conditions \cite{rezaei2015} during day and night, as well as occlusion may also affect the performance of the object detection in traffic monitoring systems \cite{gawande2020pedestrian}.

In this study, we contribute in four areas a follows:
\begin{itemize}
    \item Adapting a custom Deep Neural Network (DNN) for vehicle/pedestrian detection. 
    \item Developing an enhanced multi-object and multi-class tracking and trajectory estimation.
    \item Developing a hybrid satellite/ground-based inverse perspective mapping (SG-IPM) and calibration method for accurate localisation and distance estimation.
    \item 3D object bounding box estimation of the road users using a single-view camera.
    \item Automated short- and long-term surveillance solutions to understand traffic bottlenecks, risks, and hazards for road users. 
\end{itemize}

\noindent Figure \ref{fig-tiser} represents a sample output of our contributions including 3D detection, tracking, and environment modelling. Comprehensive details and discussions will be provided in the next sections as follows:

In Section~\ref{related} a literature review on both conventional and modern related works is conducted. 
Section~\ref{method1} introduces our methodology as an enhanced object detection and tracking algorithm followed by presenting a novel satellite/ground-based auto-calibration technique. In this section, we provide an environment modelling technique as well as the 3D representation of detected objects. Experimental results, evaluations, and comparisons with state-of-the-art %and evaluations 
will be discussed in Section~\ref{experiments}, and finally,  Section~\ref{conc} concludes the article by discussing the challenges and potentials for future works.

%=========================================
\section{Related Work}\label{related}
In this section, we review three types of related works to automated traffic surveillance systems (ATMS) including classic object detection methods, modern object detection research directions, and also the CCTV camera calibration solutions, as the prerequisite of 
any object detection methodology in the context of %any 
traffic surveillance. % tasks, prior to start any object detection.
Both classical and machine learning-based methods for automated video surveillance (AVS), automated traffic surveillance systems (ATMS), as well as the camera calibration techniques will be reviewed.

% =============== Classic
Among classical methods, a series of studies have focused on background subtraction (BGS) techniques for detecting moving objects. 
Cheung et al. \cite{Cheung2004} have compared the performance of different BGS methods such as the Mixture of Gaussian (MOG), Median filter (MF), Kalman filter (KF) and frame differentiation (FD) in various weather conditions on a road-side surveillance camera. They reported a higher precision rate using the MOG method. This method estimates various Gaussian distributions that match with the intensity distribution of the image background's content. 

Zhao et al. \cite{Zhou2007} have introduced an adaptive background estimation technique. They divide the image into small none-overlapped blocks followed by the principal component analysis (PCA) on each block's feature. Then they utilise the support vector machine (SVM) to classify the vehicles.
The method seems to be robust in partial occlusion and bad illumination conditions. However, it fails to detect stationary objects. The presented system is only evaluated on ideal highway images and neglects the crowded urban roads.

Chintalacheruvu et al. \cite{chintalacheruvu2012video} have introduced a vehicle detection and tracking system based on the Harris-Stephen corner detector algorithm. The method focuses on speed violation detection, congestion detection, vehicle counting, and average speed estimation in regions of interest. 
However, the presented method requires prior road information such as the number of lanes and road directions.  

In another approach, Cheon et al. \cite{Cheon2012} have presented a vehicle detection system using histogram of oriented gradients (HOG) considering the shadow of the vehicles to localise them. They have also used an auxiliary feature vector to enhance the vehicle classification and to find the areas with a high risk of accidents. However, the method leads to erroneous vehicle localisation during the night or day-time where the vehicles' shadows are too long and not presenting the exact location and size of the vehicle.

Although most of the discussed methods perform well in simple and controlled environments, they fail to propose accurate performance in complex and crowded scenarios. Furthermore, they are unable to perform multi-class classifications and can not distinguish between various categories of moving objects such as pedestrians, cars, buses, trucks, etc.

% =============== Modern 2D
With the emergence of \textit{deep neural networks (DNNs)}, the machine learning domain received more attention in the object detection domain. In modern object detection algorithms, Convolutional Neural Networks (CNN) learns complex features during the training phase, aiming to elaborate and understand the contents of the image. This normally leads to improvement in detection accuracy compared to classical image processing methods \cite{zou2019object}. Such object detectors are mostly divided into two categories of single-stage (dense prediction) and two-stage (sparse prediction) detectors.  
The two-stage object detectors such as RCNN family, consist of a region proposal stage and a classification stage \cite{Jiao2019}; while the single-stage object detectors such as Single-Shot Multi-Box Detector (SSD) \cite{liu2016ssd}, and You Only Look Once (YOLO) see the detection process as a regression problem, thus provide a single-unit localisation and classification architecture \cite{Jiao2019}.

Arinaldi et al.\cite{ARINALDI2018259} reached a better vehicle detection performance using Faster-RCNN compared to a combination of MOG and SVM models.

Peppa et al. \cite{Peppa2021}, developed a statistical-based model, a random forest method, and an LTSM  to predict the traffic volume for the upcoming 30 minutes, to compensate for lack of accurate information in extreme weather conditions.

Some researchers such as Bui et al. \cite{Bui2020}, utilised single-stage object detection algorithms including he YOLOv3 model for automated vehicle detection. They designed a multi-class distinguished-region tracking method to overcome the occlusion problem and lighting effects for traffic flow analysis.

In \cite{mandal2020artificial}, Mandal et al. have proposed an anomaly detection system and compared the performance of different object detection including Faster-RCNN, Mask-RCNN and YOLO. Among the evaluated models, YOLOv4 gained the highest detection accuracy. However, they have presented a pixel-based (pixel per second) vehicle velocity estimation that is not very accurate.

% =============== Modern 3D
On the other hand, the advancement of stereo vision sensors and 3D imaging has led to more accurate solutions for traffic monitoring as well as depth and speed estimation for road users. Consequently, this enables the researchers to distinguish the scene background from the foreground objects, and measure the objects' size, volume, and spatial dimensions \cite{arnold2019survey}.

LiDAR sensors and 3D point cloud data offers a new mean for traffic monitoring. In \cite{Zhang2019}, Zhang et al. have presented a real-time vehicle detector and tracking algorithm without bounding box estimation, and by clustering the point cloud space. 
Moreover, they used the adjacent frame fusion technique to improve the detection of vehicles occluded by other vehicles on the road infrastructures.  

Authors in \cite{Zhang2020}, proposed a centroid-based tracking method and a refining module to track vehicles and improve the speed estimations. Song, Yongchao, et al. \cite{song2020automatic} proposed a framework which uses binocular cameras to detect road, pedestrians and vehicles in traffic scenes.

In another multi-modal research, thermal sensor data is fused with the RGB camera sensor, leading to a noise-resistant technique for traffic monitoring \cite{Alld16contx}.

Although many studies are conducting different sensors to perform 3D object detection such as in \cite{fernandes2021point}, \cite{zhou2021rgb}, the cost of applying such methods in large and crowded cities could be significant. 
Since there are many surveillance infrastructures already installed in urban areas and there are more data available for this purpose, 2D object detection on images has gained a lot of attention in a more practical way. 

Many studies including deep learning-based methods \cite{laga2019survey}, \cite{xie2016deep3d}, have tried to utilise multi-camera and sensors to compensate for the missing depth information in the monocular CCTV cameras, to estimate the position and speed of the object, as well as 3D bounding box representation from a 2D perspective images \cite{bhoi2019monocular}. 

% =============== Calibration
Regardless of the object detection methodology, the CCTV \textit{camera calibration} is a key requirement of 2D or 3D traffic condition analysis prior to starting any object detection operation. A camera transforms the 3D world scene into a 2D perspective image based on the camera intrinsic and extrinsic parameters. Knowing these parameters is crucial for an accurate inverse perspective mapping, distance estimation, and vehicle speed estimation \cite{rezaei2017computer}. 

In many cases especially when dealing with a large network of CCTV cameras in urban areas, these parameters can be unknown or different to each other due to different mounting setups and different types of cameras. Individual calibration of all CCTVs in metropolitan cities and urban areas with thousands of cameras is a very cumbersome and costly task. Some of the existing studies have proposed camera calibration techniques in order to estimate these parameters, hence estimating an inverse perspective mapping. 

Dubska et al. \cite{Dubska2015} extract vanishing points that are parallel and orthogonal to the road in a road-side surveillance camera image, using the moving trajectory of the detected cars and Hough line transform algorithm.  This can help to automatically calibrate the camera for traffic monitoring purposes, despite low accuracy of the Hough transform algorithm in challenging lighting and noisy conditions.  

Authors in \cite{Sochor2017}, proposed a Faster-RCNN model to detect vehicles and consider car edgelets to extract perpendicular vanishing points to the road to improve the automatic calibration of the camera.  

Song et al. \cite{song20193d}, have utilised an SSD object detector to detect cars and extract spatial features from the content of bounding boxes using optical flow to track them. 
They calculate two vanishing points using the moving trajectory of vehicles in order to automatically calibrate the camera. Then, they consider a fixed average length, width and height of cars to draw 3D bounding boxes. 

However, all of the aforementioned calibration methods assume 1) The road has zero curvature which is not the case in real-world scenarios and 2) Vanishing points are based on straight-line roads. i.e. the model does not work on intersections. 

In a different approach, Kim et al. \cite{Kim2009}, consider 6 and 7 corresponding coordinates in a road-side camera image and an image with a perpendicular view of the same scene (such as near-vertical satellite image) to automatically calibrate the camera.
They introduced a revised version of RANSAC model, called Noisy-RANSAC to efficiently work with at least 6 or 7 corresponding points produced by feature matching methods.
% ربط نویزی  و افیشنتای چیست؟ این دو کلمه ضد همم هستند
However, the method is not evaluated on real-world and complex scenarios in which the road is congested and occluded by various types of road users.

Among the reviewed literature most of the studies have not investigated various categories of road users such as pedestrians and different types of vehicles that may exist in the scene. 
Moreover, there are limited researches addressing full/partial occlusion challenges in the congested and noisy environment of urban areas.
It is also notable that the performance of the latest object detection algorithm to date is not %covered 
evaluated by the traffic monitoring related researches.

%Also, 
Furthermore, very limited research has been conducted on short and long-term spatio-temporal video analysis to automatically understand the interaction of vehicles and pedestrians and their effects on the traffic flow, congestion, hazards, or accidents. 

In this article, we will aim at proving an efficient and estate-of-the-art traffic monitoring solution to tackle some of above-mentioned research gaps and weaknesses  
%efficiently monitor the traffic 
in congested urban areas. 

%=========================================
\section{Methodology}\label{method1}
We represent our methodology in four hierarchical subsections. 
In section \ref{det-section} as the first contribution, a customised and highly accurate vehicle and pedestrian detection model will be introduced. In Section \ref{track-section} and as the second contribution we elaborate our multi-object and multi-class tracker (MOMCT). Next, in Section \ref{method2}, a novel auto-calibration technique (named SG-IPM) is developed. Last but not the least, in section \ref{Env-section} we develop a hybrid methodology for road and traffic environment modelling which leads to 3D detection and representation of all vehicles and pedestrians in a road scene using a single CCTV camera. 

\begin{figure}[t!]
\centering
\includegraphics[width = 0.9\linewidth]{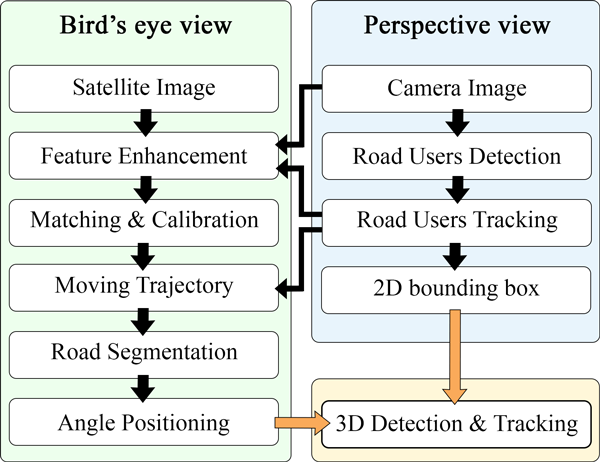}
\caption{The overall structure of the proposed methodology}
\label{fig-method}
\end{figure}

Figure \ref{fig-method} summarises the overall flowchart of the methodology, starting with a 2D camera image and satellite image as the main inputs which ultimately lead to a 3D road-users detection and tracking.

%>>>>>>>>>>>>>>>>>>>>>>>>>>>
\subsection{Object Detection and Localisation}\label{det-section}

According to the reviewed literature, the YOLO family has been proven to be faster and also very accurate compared to most of the state-of-the-art object detectors~\cite{glenn_jocher_2021_4679653}. 

We hypothesis that recent versions of the YOLO family can provide a balanced trade-off between the speed and accuracy of our traffic surveillance application. In this section we conduct a domain adaptation and transfer learning of YOLOv5. 
The Microsoft COCO dataset \cite{lin2015microsoft} consists of 80 annotated categories of the most common indoor and outdoor objects in daily life. We use pre-trained feature extraction matrices of YOLOv5 model on COCO dataset as the initial weights to train our customised model.

Our adapted model is designed to detect 11 categories of traffic-related objects which also match the MIO-TCD traffic monitoring dataset \cite{Luo2018}. 
These categories consist of a pedestrian class and 10 types of vehicles, including articulated truck, bicycle, bus, car, motorcycle, motorised vehicles, non-motorised vehicles, pickup truck, single-unit truck and work van. 
Because of the different number of classes in two datasets, the last layers of the model (the output layers) do not have the same shape to copy. Therefore, these layers will be initialised with random weights (using 100100 seeds). 
After the initialisation process, the entire model will be trained on the MIO-TCD dataset.  %This would 
We expect this would ultimately lead to a more accurate and customised model for our application.

\begin{figure*}[t!]
\centering
\includegraphics[width = 1\linewidth]{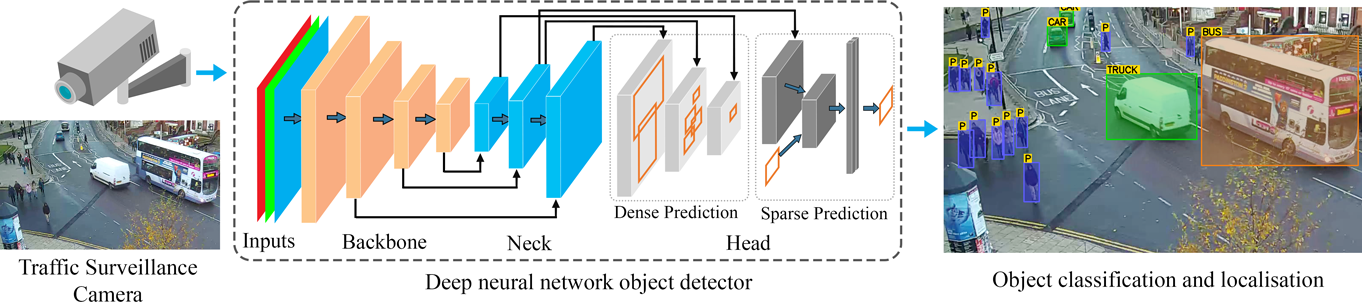}
\vspace{-5mm}
\caption{Summarised structure of the dense (single-stage) and sparse (two-stage) object detection architecture, applied to a road-side surveillance video.}
\label{fig-detection}
\end{figure*}

As shown in Figure \ref{fig-detection} the architecture of modern YOLO frameworks consists of a \textit{backbone}, the \textit{neck}, and the \textit{head}. 

The backbone includes stacked convolutional layers turning the input into the feature space. In the backbone, the Cross-Stage Partial network (CSP) \cite{wang2020cspnet} conducts shortcut connections between layers to improve the speed and accuracy of feature extractors. 

The neck consists of feature enhancers such as Spacial Pyramid Pulling (SPP) \cite{HUANG2020241}, and Path Aggregation Network (PAN) \cite{Liu2018pan}, concatenating the features extracted from initial layers (closer to the input of the model) with the end layers (closer to the head of the model) to enhance the semantic and spacial information and to improve the detection accuracy for small objects. 

The head part of YOLOv5 performs convolutional operations on the enhanced features to generate outputs (predictions) with different scales. The count and size of the outputs vary based on the number of pre-defined multi-scale grid cells, anchor boxes, and also the number of ground truth classes.
The anchor boxes are determined based on scales of the existing bounding boxes in the ground truth using the k-means clustering algorithm.
The model optimises Focal-Loss \cite{Lin2017} and Distance-IoU loss \cite{zheng2020distance} functions to classify and localise the objects, during the training process.

The latest version of YOLOv5 (to the date of this article), incorporates $4$ head outputs with the strides of $8$, $16$, $32$ and $64$. The output scale with the stride of $64$ is added to improve the detection accuracy of relatively large objects in multi-faceted and comprehensive datasets. However, in most traffic monitoring scenes, including this research, the cameras are placed at a height of at least 3 metres from the ground and with a distance of more than 5 metres from the objects of interest (road users). This means the cameras hardly includes any extra-large objects that can fill up the entire image plane. 

Therefore, we consider $3$ different stride scales of $8$, $16$, $32$ for the head part of our model and the $k$-means algorithm with $9$ cluster centroids, yielding to $3$ anchor boxes for each grid-cell. 
This means the model can detect $3$ objects in each grid cell.
Our preliminary evaluations confirm accuracy improvements by using the $3$ head scales rather than the $4$ head scales (more details in Section \ref{obj-section}).
The model predicts offset of bounding boxes with respect to corresponding anchor box in each grid cell.
Assuming $(x_c,y_c)$ as the top-left corner of the grid-cell, $h_{a}$ and $w_{a}$ as the height and width of an anchor box in that grid cell, a bounding box with the centre $(x_{b},y_{b})$, height $h_{b}$ and width $w_{b}$ is calculated by the predicted offset $(x_{o},y_{o},w_{o},h_{o})$ as follows:
\begin{equation}
\begin{matrix}
x_b = \sigma{(x_{o})} + x_{c} \\
y_b = \sigma{(y_{o})} + y_{c} \\
w_b = w_{a} \times e^{w_{o}}  \\
h_b = h_{a} \times e^{h_{o}}
\end{matrix}
\end{equation}
\noindent where $\sigma$ is the Sigmoid function, normalising the input between $0$ and $1$. 

Eventually, the model produces a set $\mbox{D}$ for each image witch contains $(x_{{b}},y_{b},w_{{b}}, h_{b}, \mathfrak{s} ,\textbf{c})$ for each object, where $\mathfrak{s} $ is objectness confidence score and $\textbf{c}$ is a vector of classification probabilities with a length equal to the number of classes.

We consider the coordinates of the middle point at the bottom side of each bounding box as the reference point of the detected objects.
This is the closest contact point of the vehicles and pedestrians to the ground (the road surface):
 \begin{equation}\label{all_locations}
(\hat{x},\hat{y}) = (x_b,y_b+\frac{h_b}{2})
\end{equation}

%>>>>>>>>>>>>>>>>>>>>>>>>>>>

\begin{figure*}[t!]
\centering
\includegraphics[width = 1\linewidth]{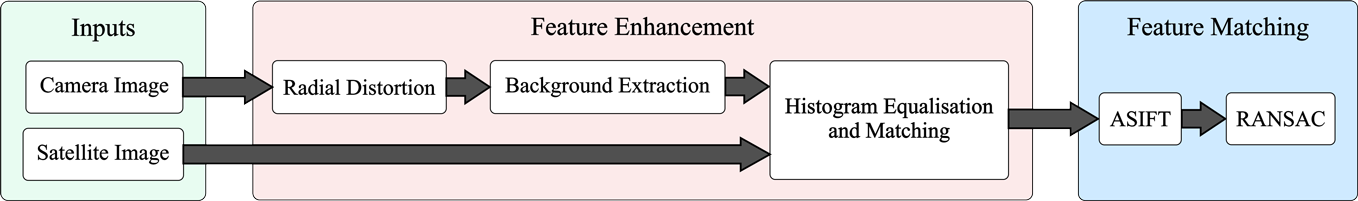}
\vspace{-1mm}
\caption{The hierarchical structure of the proposed feature matching model.}
\label{fig-matcher}
\end{figure*}

\subsection{Object Tracking and Moving Trajectory}\label{track-section}
DeepSORT \cite{Wojke2017} is a common DNN-based object tracking algorithm that extracts appearance features to track the detected objects. However, it comes with a comparatively high computational cost which is a negative point for multi-modal applications such as our traffic surveillance application.

Therefore, we aim at enhancing the Simple Object Real-time Tracking (SORT) algorithm \cite{Bewley_2016} by %perform
developing a fast tracking model called multi-object and multi-class tracker (MOMCT), while maintaining a high level of tracking accuracy.

The SORT algorithm assigns a unique ID to each object by computing the Intersection over Union (IoU) between detected bounding boxes in consequent frames of the input video. However, this process is only applicable to a single class and each class needs to be dealt with separately. As a result, in some cases, the object detector assigns a new class to an object (bounding box) which is not aligned with the SORT object tracker estimation. In such cases, the tracker sees it as a new object, assigns a new ID to it and consequently loses the previous tracking. 
 
To overcome this issue, we integrate a category vector $\acute{\textbf{c}} \in \mathbb{W}^{1\times 11}$ for 11 categories of detected objects in the internal Kalman filter of the SORT tracker. The category vector is the one-hot encoded representation of the detected class vector $\textbf{c}$, in which the highest class probability is shown by $1$ and the rest of the probabilities by $0$. 

Exploiting the smoothing effect of the Kalman filter would filter out the bouncing of detected categories through the sequence of frames. 
Also, it enables the SORT to calculate IoU between the bounding boxes of different categories. This yields a multi-object and multi-category ID assignment.

The state matrix of the new Kalman filter is defined as follows:
\begin{equation}\label{Kstate}
\grave{\textbf{x}} = [\hspace{0.1cm}\hat{x} \hspace{0.2cm} \hat{y} \hspace{0.2cm} s_b \hspace{0.2cm} r_b \hspace{0.2cm} \dot{x} \hspace{0.2cm} \dot{y}\hspace{0.2cm} \dot{s} \hspace{0.2cm}|\hspace{0.1cm}\acute{\textbf{c}}\hspace{0.1cm}]^T
\end{equation}
\noindent where $s_b={w_b}\times {h_b}$ denotes the scale (area), $r_b$ is the aspect ratio, $\dot{x}$, $\dot{y}$ and $\dot{s}$ are the velocities of $\hat{x}$, $\hat{y}$ and $s_b$, respectively. Similarly, we represent the observation matrix of the revised Kalman filter as follows:

\begin{equation}
\label{Kobserve}
\grave{\textbf{z}} = [\hspace{0.1cm}\hat{x} \hspace{0.2cm} \hat{y} \hspace{0.2cm} s_b \hspace{0.2cm} r_b \hspace{0.2cm} | \hspace{0.1cm} \acute{\textbf{c}}\hspace{0.1cm}]^T
\end{equation}
In order to determine the trajectory of objects, we introduce two sets of $V$ and $P$ as the tracker-ID of detected vehicles and pedestrians, respectively.

The trajectory set of each vehicle $(v_i)$ and pedestrian $(p_i)$ can be calculated based on temporal image frames as follows:

\begin{equation}\label{traj}
\begin{aligned}
M_{v_i} = \{ (\hat{x}_{v_i}^t, \hat{y}_{v_i}^t) \hspace{0.2cm}:\hspace{0.2cm} \forall t \in T_{v_i} \} \\
M_{p_i} = \{ (\hat{x}_{p_i}^t, \hat{y}_{p_i}^t) \hspace{0.2cm}:\hspace{0.2cm} \forall t \in T_{p_i} \}
\end{aligned}
\end{equation}

\noindent where $T_{v_i}$ and  $T_{p_i}$ are the sets of frame-IDs of the vehicles $v_i$ and pedestrians $p_i$ and $(\hat{x}^t, \hat{y}^t)$ is the location of the object $v_i$ or  $p_i$ at frame $t$.

Finally, moving trajectories of all tracked objects are defined as the following sets:
\begin{equation}\label{traj2}
\begin{aligned}
M_{V} = \{ M_{v_i} \hspace{0.2cm}:\hspace{0.2cm} \forall v_i \in V \} \\
M_{P} = \{ M_{p_i} \hspace{0.2cm}:\hspace{0.2cm} \forall p_i \in P \}
\end{aligned}
\end{equation}

%>>>>>>>>>>>>>>>>>>>>>>>>>>>
\subsection{Camera Auto-calibration} \label{method2}
The intuition behind this part of the study is to apply an automatic IPM camera calibration setup where and when no information about the camera and mounting specifications are available. This makes our study applicable for most CCTV traffic surveillance cameras in city roads and urban areas, as well as other similar applications, without the requirements of knowing the camera intrinsic parameters, height and angle of the camera. 

We exploit a top-view satellite image from the same location of the CCTV camera and develop a hybrid satellite-ground based inverse perspective mapping (SG-IPM) to automatically calibrate the surveillance cameras.
This is an end-to-end technique to estimate the planar transformation matrix $\textbf{G}$ as per Equation \ref{3by3t} in Appendix A.
The matrix $\textbf{G}$ is used to transform the camera perspective image to a bird's eye view image.

Let's assume $(x,y)$ as a pixel in a digital image container $\textbf{I}: \mathcal{U} \rightarrow [0, 255]^3$  were $\mathcal{U} = [[0; w-1] \times [0; h-1]]$ represents the range of pixel locations in a 3 channel image, and $w$, $h$ are width and height of the image. 

Using ( $\hat{ }$ ) to denote the perspective space (i.e. camera view), and, ( $\check{ }$ ) for inverse perspective space, 
we represent the surveillance camera image as $\hat{\textbf{I}}$, the satellite image as $\grave{\textbf{I}}$, and bird's eye view image as $\check{\textbf{I}}$ which is calculated by the linear transformation $\textbf{G}: \hat{\textbf{I}} \rightarrow \check{\textbf{I}}$.

Since the coordinates of the bird's eye view image approximately matches the satellite image coordinates (i.e. ($\grave{\textbf{I}} \approx \check{\textbf{I}}$), the utilisation of the transformation function  $(\check{x},\check{y}) = \Lambda((\hat{x},\hat{y}),\textbf{G})$ (as defined in Appendix A) would transform the pixel locations of $\hat{\textbf{I}}$ to the $\grave{\textbf{I}}$.
Similarly, $\textbf{G}^{-1}$ inverts the mapping process. In other words,  $(\hat{x},\hat{y}) = \Lambda((\check{x},\check{y}),\textbf{G}^{-1})$ transforms the pixel locations from $\check{\textbf{I}}$ to the $\hat{\textbf{I}}$.

In order to solve the linear equation \ref{3by3t}, at least four pairs of corresponding points in $\hat{\textbf{I}}$ and $\grave{\textbf{I}}$ are required. Therefore, we would need to extract and match similar features pairs from both images.
These feature points should be robust and invariant to rotation, translation, scale, tilt, and also partial occlusion in case of high affine variations.

Figure \ref{fig-matcher} represents the general flowchart of our SG-IPM technique, which is fully explained in the following sub-sections, including \textit{feature enhancement} and \textit{feature matching}:\\

\subsubsection{Feature Enhancement}

Three types of feature enhancement are addressed before applying the calibration processes:

\begin{itemize}
    \item Radial distortion correction
    \item Background removal
    \item Histogram matching
\end{itemize}

\noindent \textbf{\textit{Radial Distortion}}: Some of the road-side cameras have non-linear radial distortion due to their wide-angle lens which will affect the accuracy of the calibration process and monitoring system to estimate the location of the objects. 

Such type of noise would also reduce the resemblance between $\hat{\textbf{I}}$ and $\grave{\textbf{I}}$ images, especially, in the case that we want to find similar feature points.

\begin{figure}[t!]
\includegraphics[width = 1\linewidth]{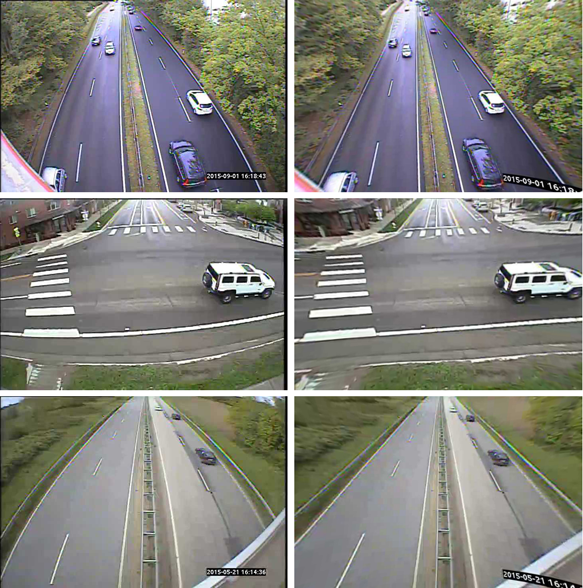}
\caption{Eliminating the radial distortion in MIO-TCD dataset samples \cite{Luo2018}. 
Left column: original images. Right column: %corrected 
rectified images after barrel distortion removal.
}\label{RD}
\end{figure}

Examples of the barrel-shaped radial noise are shown in Figure \ref{RD}, left column. 
Similar to a study by Dubská et al \cite{Dubska2015}, we assume the vehicles traverse between the lanes in a straight line. 
We use the vehicles' trajectory sets to remove radial distortion noise.
For each vehicle $v_i$, a polynomial radial distortion model is applied to the location coordinates $(\hat{x}_{v_i},\hat{y}_{v_i})$ of the vehicle's trajectory set ($M_{v_i}$) as follows:

\begin{equation}\label{Radial1}
\begin{aligned}
(\bar{x},\bar{y}) = 
((\hat{x}_{v_i}-x_s)(1+k_1r^2+k_2r^4+...), \\
 (\hat{y}_{v_i}-y_s)(1+k_1r^2+k_2r^4+...)) \\
\end{aligned}
\end{equation}
\begin{equation}\label{Radial3}
r = \sqrt{(\hat{x}_{v_i}-x_s)^2 + (\hat{y}_{v_i} -y_s)^2}
\end{equation}

\noindent where $(\bar{x} , \bar{y})$ is the corrected location of the vehicle, $(x_s,y_s)$ denotes the centre of the radial noise, $K=\{ k_1,k_2,...\}$ are the unknown scalar parameters of the model which need to be estimated, and $r$ is the radius of the distortion with respect to the centre of the image.

A rough estimation of $k_1$ and $k_2$ would be sufficient to remove the major effects of such noise. 
To this regard, each point of the moving trajectories would be applied to the Equation~\ref{Radial1} yielding to transformed trajectory set $\bar{M}_{v_i}$.
Then, the optimal values of $k_1$ and $k_2$ would be achieved by minimising the sum of squared errors between the best fitting line $\ell$ to the ${M}_{v_i}$ and $\bar{M}_{v_i}$ as follows:
\begin{equation}\label{Radial4}
K = \arg \hspace{0.1cm} \min _k\sum\limits_{{v_i} \in V} \hspace{0.1 cm} \sum\limits_{\bar{l}_j \in \bar{M}_{v_i}} (\ell.\bar{l}_j)^2
\end{equation}
where $\bar{l}_j$ is the corrected pixel location of the vehicle $v_i$ belonging to the transformed moving trajectory set $\bar{M}_{v_i}$.

Finally, the optimal parameters will be estimated using $(1+\lambda)$-ES evolutionary algorithm with $\lambda = 8$ as discussed in \cite{Dubska2015}.\\

\noindent \textbf{\textit{Background Extraction}}: Since $\hat{\textbf{I}}$ and $\grave{\textbf{I}}$ images are captured using two different cameras (ground camera vs. aerial satellite camera) and in different dates, times, or weather conditions, the images may seem inconsistent and different. This is mostly due to the existence of different foreground objects and road users on each image. This makes it hard to find analogous features to match. 

To cope with that challenge, we extract the background of image $\hat{\textbf{I}}$ by eliminating the moving objects.
We apply an accumulative weighted sum over the intensity value for a period of $\mathsf{n}_t$ (frames) to remove the effect of the temporal pixel value changes as follows:

\begin{equation}\label{accsum}
 \hat{\textbf{B}}^t=
 (1-\alpha) \hat{\textbf{B}}^{t-1} + (\alpha \hspace{0.1cm} \hat{\textbf{I}}^{t}) \hspace{.3cm}  ,\hspace{.3cm} 1 \leq t \leq \mathsf{n}_t
 \end{equation}

\noindent where initially $\hat{\textbf{B}}$ is the accumulative variable 
and $\hat{\textbf{B}}^0$ is equal to the first input frame $\textbf{I}^0$, $\alpha$ is the weighted coefficient that determines the importance of the next incoming frame. 
Our experiment shows that $\alpha = 0.01$, and $\mathsf{n}_t \approx 70$ frames is usually sufficient to remove the foreground objects in most urban and city roads with a moderate traffic flow.

\begin{figure}[t!]
\vspace{-4mm}
\centering
\subfloat[Laidlaw Library CCD original video footage]{
\resizebox{4.0cm}{!}{\includegraphics{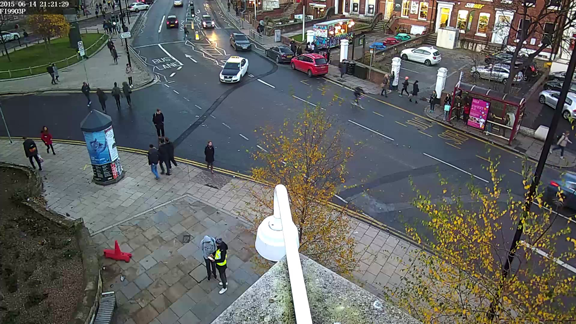}\label{be_leeds}}}\hspace{1pt}
\subfloat[The generated Laidlaw Library background video]{
\resizebox{4.0cm}{!}{\includegraphics{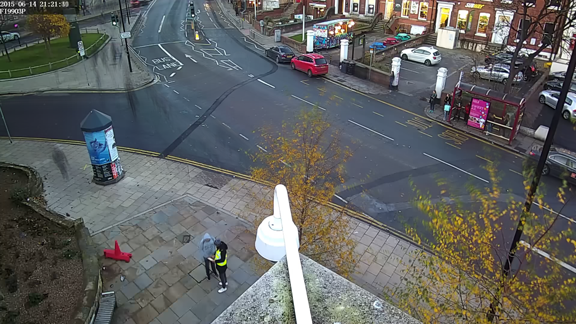}\label{be_leeds_b}}}\vspace{1pt}

\subfloat[GRAM original video footage]{
\resizebox{4.0cm}{!}{\includegraphics{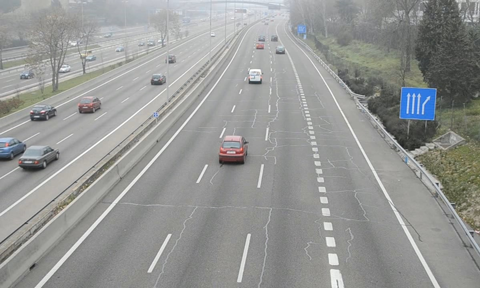}\label{be_gram}}}\hspace{1pt}
\subfloat[GRAM background video]{
\resizebox{4.0cm}{!}{\includegraphics{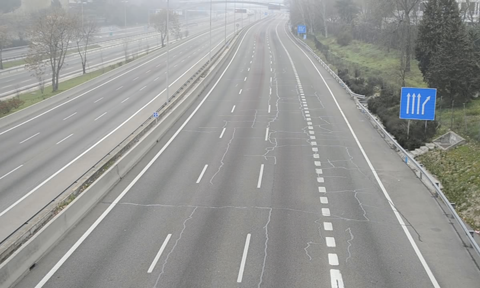}\label{be_gram_b}}}\vspace{1pt}

\subfloat[UA-DAT original video footage]{
\resizebox{4.0cm}{!}{\includegraphics{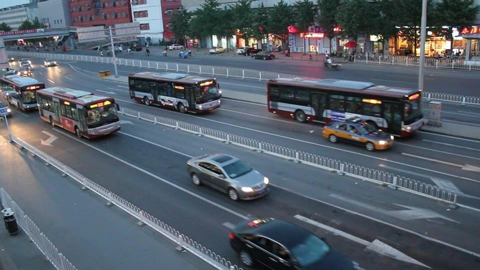}\label{be_uadet}}}\hspace{1pt}
\subfloat[UA-DAT background video]{
\resizebox{4.0cm}{!}{\includegraphics{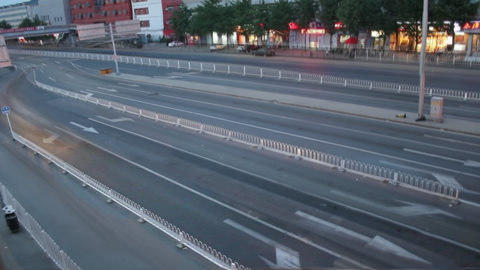}\label{be_uadet_b}}}\vspace{1pt}

\hspace{1pt}
\caption{Samples of background extraction on the UK Leeds Laidlaw Library (Parkinson Building) dataset, GRAM dataset \cite{guerrero2013iwinac}, and UA-DAT dataset \cite{UA-DETRAC2020} .}\label{BE}
\end{figure}

Figure \ref{BE} shows samples of the background extraction method applied to various roads and traffic scenarios.\\

\noindent \textbf{\textit{Histogram Matching}}: Lighting condition variation is  another barrier that makes it hard to find and match similar feature points between $\hat{\textbf{I}}$ and $\grave{\textbf{I}}$.

We utilise a colour correlation-based histogram matching \cite{Niu2018} which adjusts the hue and luminance of $\hat{\textbf{I}}$ and $\grave{\textbf{I}}$ into the same range. 
The algorithm can be extended to find a monotonic mapping between two sets of histograms.
The optimal monotonic colour mapping $E$ is calculated to minimise the distance between the two sets simultaneously.

\begin{equation}\label{histogram-eq}
\dbar = 
\arg \hspace{0.2cm} \min_{E}\sum_{i= n_p}d(E(\hat{\textbf{I}}_{i}^g),\hspace{0.1cm}\grave{\textbf{I}}_{i}^g)
\end{equation}
\noindent where $\hat{\textbf{I}}^g$ and  $\grave{\textbf{I}}^g$ are grey-level images, $n_p$ is the number of pixels in each image, 
% [AZ] Added:
$E$ is histogram matching function,
and ${\displaystyle d(\cdot ,\cdot )}$ is a 
% [AZ] Added:
Euclidean 
distance metric between two histograms.

\begin{figure}[t!]
\centering
\subfloat[Background extraction]{
\resizebox{8.7cm}{!}{\includegraphics{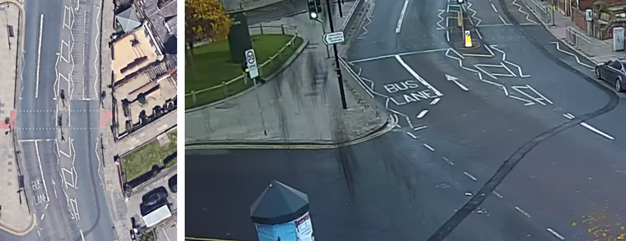}\label{backMatch}}}\vspace{1pt}
\subfloat[The outcome of histogram matching]{
\resizebox{8.7cm}{!}{\includegraphics{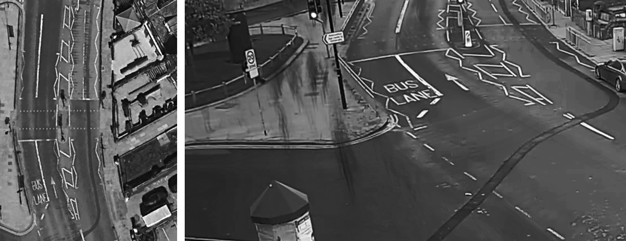}\label{colorMatch}}}\vspace{1pt}
\caption{Histogram matching algorithm applied to the Leeds University Laidlaw Library (Parkinson Building) surveillance camera (right column), and the satellite image of the same location (left column).}
\label{hma}
\end{figure}

Figures \ref{backMatch} and \ref{colorMatch} show the results of the background extraction and histogram matching process, respectively.\\

\subsubsection{Feature Matching}

To handle the high affine variation between the images, we adapt the Affine Scale-Invariant Feature Transform (ASIFT) \cite{yu2011} method. This method generates view samples along different latitude and longitude angles of the camera. Then it applies Scale-Invariant Feature Transform (SIFT) \cite{Lowe1999} algorithm. 
This makes it invariant to all parameters of the affine transformation and a good candidate to match the features between $\hat{\textbf{I}}^g $ and  $ \grave{\textbf{I}}^g$. 

However, there might be some outliers between the matching features causing inaccurate estimation of the matrix $\textbf{G}$. To remove these outliers, we use Random Sample Consensus (RANSAC) \cite{Fischler1981}, which is an iterative learning algorithm for parameter estimation. 
In each iteration, the RANSAC algorithm randomly samples four corresponding pairs among all matching points between $\hat{\textbf{I}}^g $ and  $ \grave{\textbf{I}}^g$.
Then, it calculates the $\textbf{G}$ matrix using the collected samples and performs a voting process on all matching feature-pairs in order to find the best matching samples.

Considering $\hat{l}_f $ and  $\grave{l}_f$ as the locations of matching pairs, the following criteria can be defined to evaluate the best candidates:

\begin{equation}\label{voting-cond}
\begin{matrix}
F_n = 
\begin{cases} 
 1  & d( \Lambda(\hat{l}_f , \textbf{G}), \hspace{0.2cm} \grave{l}_f) < \tau_{\mbox{z}} \\
 0 & \hspace{1.1cm} \mbox{Otherwise}
 \end{cases}
\end{matrix}
\end{equation}

\noindent where $F_n$ is the result of voting for the $n$-th pair, $\tau_{\mbox{z}}$ is a distance threshold to determine whether a pair is an inlier or not, and $d$ is the Euclidean distance measure. Consequently, the total number of inlier votes ($\hslash_i$) for the matrix $\textbf{G}$ in the $i$-th iteration will be calculated as follows:

\begin{equation}\label{voting}
  \hslash_i = \sum\limits_{n=1}^{\eta} F_n \hspace{0.2cm} ,\hspace{0.2cm}  i \in {\zeta}
\end{equation}

\noindent where $\eta$ is the total number of matching feature-pairs, and $\zeta$ is the total number of RANSAC iterations which is defined as follows:

\begin{equation}\label{iter_count}
	\zeta =  \frac{\log (1 - \rho)}{\log (1-\epsilon^\gamma)}
\end{equation}

\noindent where $\epsilon$ is the probability of a pair being inlier (total number of inliers divided by $\eta$), $\gamma$ is the minimum number of random samples ($4$ feature-pairs in our setting, which is the least requirement in order to calculate $\textbf{G}$ matrix), 
%
%[fm]commented :and $\rho$ is the probability of a successful iteration.An iteration is successful if all of the $\varsigma$ sampled pairs are inliers.
%[fm] added: 
and $\rho$ is the probability of all $\varsigma$ sampled pairs being inliers in %one 
an iteration.
After the end of the iterations, the $\textbf{G}$ matrix with the highest vote will be elected as the suitable transformation matrix between $\hat{\textbf{I}}^g $ and  $ \grave{\textbf{I}}^g$.

\begin{figure}[t!]
\centering
\subfloat[ASIFT feature matching results]{
\resizebox{8.7cm}{!}{\includegraphics{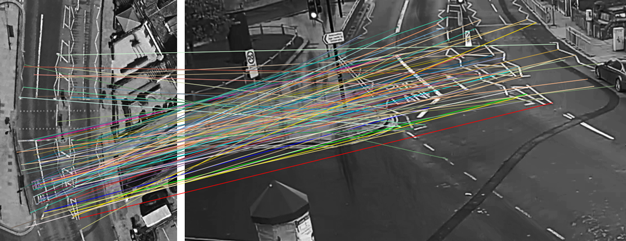}\label{initMatch}}}\vspace{1pt}
\subfloat[RANSAC inlier matches]{
\resizebox{8.7cm}{!}{\includegraphics{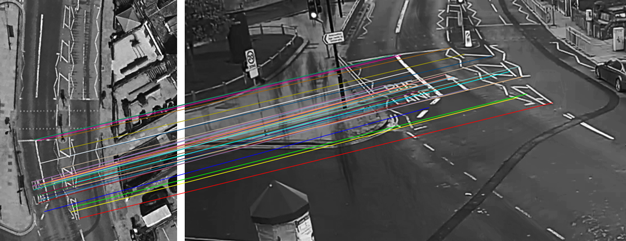}\label{bestMatch}}}%\hspace{1pt}
\caption{Feature matching process applied to the Leeds University Laidlaw Library (Parkinson Building) surveillance camera (right side) and the satellite image of the same location (left side).}
\label{fig=match}
\end{figure}

Figure \ref{fig=match} represents an example of the feature matching process applied to a real-world scenario. 
Figures \ref{initMatch} and \ref{bestMatch} show the results of the ASIFT algorithm and the RANSAC method, respectively. 

\begin{figure}[t!]
\centering
\includegraphics[width = 1\linewidth]{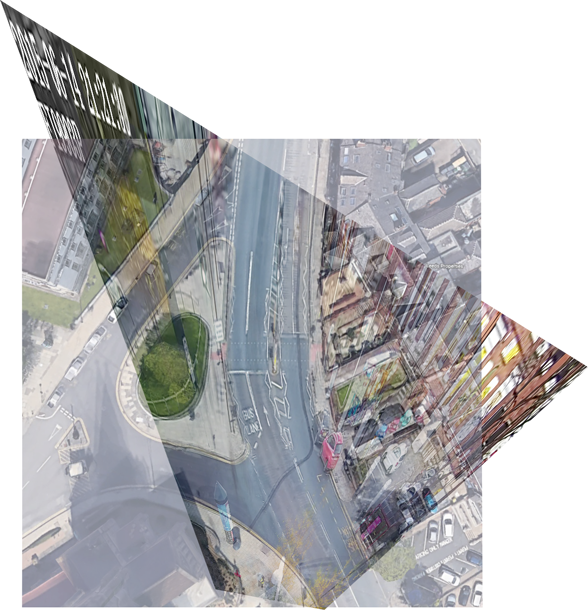}
\caption{ %Result of applying 
Overlapping the estimated BEV image $\check{\textbf{I}}$ to the ground truth satellite image $\grave{\textbf{I}}$ of the same location.}\label{fig-overlap}
\end{figure}

Eventually, we apply the $\textbf{G}$ matrix on coordinates of interest in $\hat{\textbf{I}}$ (such as positions of detected objects $(\hat{x},\hat{y})$), to estimate their corresponding coordinates in $\check{\textbf{I}}$:
\begin{equation}
(\check{x},\check{y}) = \Lambda((\hat{x},\hat{y}),\textbf{G})
\end{equation}
As it can be visually confirmed, Figure \ref{fig-overlap} shows a very accurate result of %
%added
mapping of %applying 
the estimated matrix $\textbf{G}$ on $\hat{\textbf{I}}$ coordinates. 
The resulting image has been projected on the $\grave{\textbf{I}}$ to make the intersection of overlapping areas more visible, and also easier for a visual comparison.

\subsection{3D Environment Modelling and Traffic Analysis} \label{Env-section}

Automated analysis of traffic scene videos via surveillance cameras is a non-trivial and complex task. This is mainly due to the existence of various types of objects such as trees, buildings, road users, banners, etc in various sizes and distances. Occlusion and lighting conditions are additional parameters that makes it even more complex. 
In this section, we elaborate our techniques of providing an abstract visual representation of the environment, objects of interest, traffic density, and traffic flow.
In order to achieve a 3D bounding box modelling and representation of the road users, we require to identify and recognise the following properties for the road users and the road scene:

\begin{itemize}
    \item Vehicle's velocity ($\vartheta$)
    \item Vehicle's heading angle ($\theta$)
    \item Road boundary detection 
\end{itemize}
Initially, the estimation process of the vehicle's velocity ($\vartheta$) and heading angle ($\theta$) is described. Then we apply semantic segmentation on the satellite image to extract the road's region and boundary, and finally, we propose a method to create 3D bounding boxes.

%>>>>>>>>>>>>>>>>>>>>>>>>>>>
\subsubsection{Speed Estimation}
\label{Sec.speed.method}

Assuming the location of vehicle $v_i$ in the current time as $\check{l}_{v_i}^t= (\check{x}_{v_i}^t,\check{y}_{v_i}^t)$ and in the previous time as $\check{l}_{v_i}^{t-1}= (\check{x}_{v_i}^{t-1},\check{y}_{v_i}^{t-1})$ in the trajectory set ${M}_{v_i}$, the velocity can be calculated as follows:
\begin{equation}\label{velocity}
\vartheta_{v_i} = \frac{d(\check{l}^t_{v_i},\check{l}^{t-1}_{v_i}) }{\Delta t} \times \iota
\end{equation}

\noindent where $\Delta t$ is the time difference in seconds, and $\iota$ is the length of one pixel in meters (pixel-to-meter ratio).

To calculate $\iota$, we consider a well-known measure, or an standard object, sign, or road marking %distance 
with a known size in the scene, %camera image 
such as the width of the 2-lane city roads (which is $7m$ in the UK) or the length of white lane markings (which is e.g. $3m$ in Japan) as a real-world distance reference. %criterion.
Dividing the real-distance reference by the number of the pixels in the same region of the satellite image, gives us the pixel-to-meter ratio ($\iota $).

Although the integrated Kalman filter of the object tracker in the perspective image reduces the object localisation noise to some extent, the SG-IPM method may add up some additional noise in the bird's eye view image, which in return leads to an unstable bird's eye view mapping and estimations. 
To overcome this issue, we have applied a constant acceleration Kalman filter on the object locations $(\check{x},\check{y})$ which models the motion of objects. The state matrix of this Kalman filter is defined as:
\begin{equation}
	\breve{\textbf{x}} = [\hspace{0.1cm}\check{x} \hspace{0.2cm} \check{y} \hspace{0.2cm} \dot{x} \hspace{0.2cm} \dot{y}\hspace{0.2cm} \ddot{x} \hspace{0.2cm} \ddot{y} \hspace{0.1cm}]^T
\end{equation}
\noindent where the $\dot{x}$ and $\dot{y}$ are the velocity and $\ddot{x}$ and $\ddot{y}$ are the accelerations in $\check{x}$ and $\check{y}$ directions, respectively. 

We represent the Kalman transition matrix ($\breve{\textbf{A}}$) as follows:

\begin{equation}\label{transition-materix}
\breve{\textbf{A}}=
\begin{bmatrix}
1 & 0 & t_w & 0 &\frac{t_w^2}{2}&0\\
0 & 1 & 0  & t_w & 0 & \frac{t_w^2}{2}\\
0 & 0 & 1 & 0 & t_w & 0\\
0 & 0 & 0 & 1& 0 & t_w
\end{bmatrix}
\end{equation}
\noindent where $t_w = \frac{1}{\textit{fps}}$  is the real-world time between the former and the current frame depending on the camera frame rate (frame per second, $\textit{fps}$). The observation matrix $\breve{\textbf{z}}$ can be defined as follows:
\begin{equation}
	\breve{\textbf{z}} = [\hspace{0.1cm}\check{x} \hspace{0.2cm} \check{y} \hspace{0.1cm}]^T
\end{equation}

Using the Kalman-based smoothed location $(\check{x},\check{y})$ and the frame by frame velocity of objects $\dot{x}$, the speed of a vehicle will be calculated (in \textit{mph}) as follows:
\begin{equation}
 \vartheta_{v_i} = \dot{x}_{v_i} \times \iota 
 \end{equation}

\noindent where $\dot{x}_{v_i}$ is the "pixels per second" velocity of the vehicle $v_i$,, and $\iota$ is the pixel-to-mile ratio. Samples of estimated speeds (in mph) is shown on top-left corner of the vehicle bounding boxes in Figure \ref{fig-tiser}, bottom row.

In case of missing observations due to e.g. partial occlusion, we predict the current location of vehicles using the process step of the Kalman filter ($\breve{\textbf{A}} \hspace{0.1cm} .  \breve{\textbf{x}}$) and buffering the predicted locations up to an arbitrary number of frames.% (\ref{occlusion}).
\\

%>>>>>>>>>>>>>>>>>>>>>>>>>>>
\subsubsection{Angle Estimation}
The heading angle of a vehicle can be calculated as follows:
\begin{equation}\label{ang}
\theta_{v_i} = 
\theta(\check{l}_{v_i}^t,\check{l}_{v_i}^{t-1}) =
\tan^{-1} (\frac{\check{y}_{v_i}^t-\check{y}_{v_i}^{t-1}}{\check{x}_{v_i}^t-\check{x}_{v_i}^{t-1}})
\end{equation}

The angle estimation is very sensitive to the displacement of vehicle locations, and even a small noise in localisation can lead to a significant change in the heading angle.
However, in the real world the heading angle of vehicles would not change significantly in a very short period of time (e.g. between two consequent frames). 

\begin{figure}
\centering
\includegraphics[width = 0.8\linewidth]{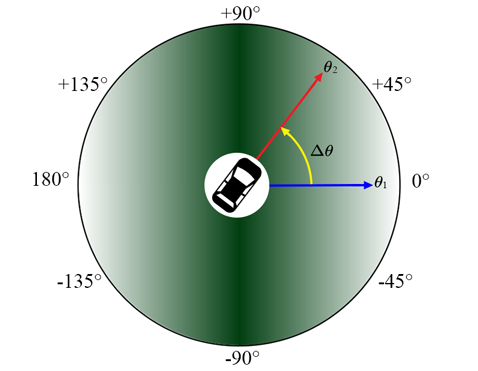}
% edited
%\caption{Smoothing sudden angle changes ($\Delta{\theta}$) between two consequent frames.}
\caption{$\Delta{\theta}$ cosine suppression operation. The darker zones receive a lower coefficients which in turn suppress any large and sudden angular changes between two consequent frames.}
\label{fig-angleBin}
\end{figure}

We introduce a simple yet efficient Angle Bounce Filtering (ABF) method to restrict sudden erroneous angle changes between the current and previous angle of the vehicle: 

\begin{equation}
 \Delta \theta_{v_i} =\theta_{v_i}^{\hspace{0.03 cm}t}-\theta_{v_i}^{\hspace{0.03 cm}t-1}
\end{equation}

\noindent where $\Delta \theta_{v_i}$ is  in the range of $[-180^\circ, 180^\circ]$.
In order to suppress high rates of the changes, we consider a cosine weight coefficient ($\mathsf{w}$) as follows:
\begin{equation}
\mathsf{w} = \frac{\cos ((4\pi  \times \tilde{\Delta}) +1)}{2 }
\end{equation}

\noindent where $\tilde{\Delta}$ is the normalised value of $\Delta \theta_{v_i}$ within the range of $[0, 1]$. The coefficient yields to "0" when the $\Delta \theta_{v_i}$ approaches to $\pm 90^\circ$ to neutralise the sudden angle changes of the vehicle. Similarly, the coefficient yields to "1" when the $\Delta \theta_{v_i}$ approaches to $0 ^\circ$ or $\pm 180 ^\circ$ to maintain the natural forward and backward movement of the vehicle. Figure \ref{fig-angleBin} illustrates the smoothed values of $\mathsf{w}$ by green colour spectrum. The darker green, the lower the coefficient.

Finally, we rectify the vehicle-angle as follows:
\begin{equation}
\tilde{\theta}_{v_i}^{\hspace{0.03 cm}t} = 
\theta_{v_i}^{\hspace{0.03 cm}t-1} + (\mathsf{w}\hspace{0.1cm} \times \Delta  \theta_{v_i}\hspace{0.1cm})
\end{equation}

In some cases the moving trajectory may not be available; for instance, when a vehicle appear on the road-scene for the first time or some vehicles are stationary (or parked) during their entire presence in the scene. For such cases the heading direction of the vehicle cannot be directly estimated as no prior data is available about the vehicle movement history. However, we can still calculate the angle of the vehicles by calculating a perpendicular line from the vehicle position to the closest boundary of the road. Identifying the border of the road requires a further road segmentation operation. 

%>>>>>>>>>>>>>>>>>>>>>>>>>>>
%\subsubsection{Road Segmentation}

Some of the existing deep-learning based studies such as \cite{wu2019towards} mainly focus on segmenting satellite imagery which are captured from a very high altitude and heights comparing to the height of CCTV surveillance cameras. 

Moreover, there are no annotated data available for such heights to train a deep-learning based road segmentation model.
In order to cope with that limitation, we adapt a Seeded Region Growing method (SRG) \cite{adams1994seeded} on intensity values of the image $\check{\textbf{I}}$.

We consider the moving trajectory of vehicles traversing the road in $\check{\textbf{I}}$ domain ($\Lambda(M_{v_i}, \textbf{G}) \hspace{0.15 cm} \forall v_i \in V$), as initial seeds for the SRG algorithm.
In the first step, the algorithm calculates the intensity difference between each seed and its adjacent pixels. 
Next, the pixels with an intensity distance less than a threshold $\tau_{\alpha}$, are considered as connected regions to the seeds.
Utilising these pixels as new seeds, the algorithm repeats the above steps until no more connected regions are found. 
At the end of the iterations, the connected regions represent the segment of the road.

\begin{figure}[t!]
\centering
\includegraphics[width = 0.8\linewidth]{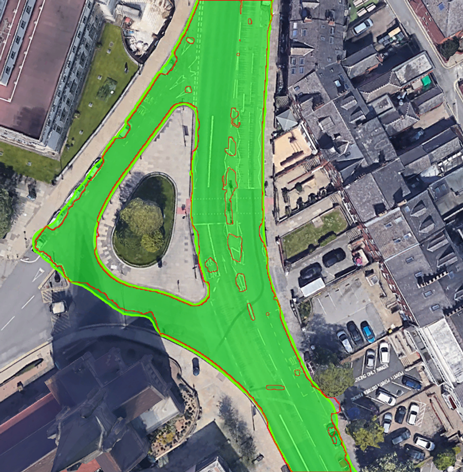}
\caption{Road segmentation on the satellite image. The red lines represent the initial segmentation result extracted from the SRG method, and the green region is the final segmentation output after applying the morphological operations.
} \label{fig-satseg}
\end{figure}

Due to a large intensity variations among adjacent pixels in the road segment (such as white lane markings vs the dark grey asphalt coatings), there might be %unsegmented gaps between the segmented areas 
some fragmented road boundary segments 
as shown in Figure \ref{fig-satseg} (the regions denoted by red lines at the centre and around the road).

We apply morphological dilation operations with $3 \times 3$ kernel size, to expand the segmented area and fill these small gap regions. Also, an erosion operation with the same kernel size is performed to smooth the road region by removing the sharp edges and spikes of the road boundaries.  Figure \ref{fig-satseg}, green regions, represent the segmentation results.

The road segmentation process can be done as an offline procedure before the real-time traffic monitoring operation starts. The scene need to be monitored until sufficient vehicle locations (seeds) are detected to segment the entire road region. Since the initial seeds are moving trajectories of vehicles, the monitoring time may vary for different scenes depending on the presence of the vehicles traversing the road. 
Based on our experience this may vary from 5 seconds to 5 minutes depending on the live traffic flow. 

\begin{figure}[t!]
\centering
\includegraphics[width = 0.8\linewidth]{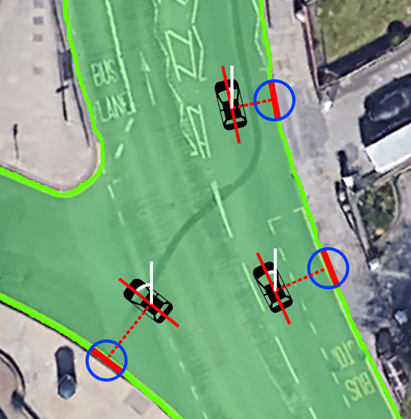}
\caption{Reference angle estimation process with respect to nearest road boundary. The boundaries are denoted with green lines, which are extracted by application of the Canny edge detector on the road segment borders.}
\label{fig-satsegang}
\end{figure}

In order to calculate the reference heading angle for each vehicle ($v_i$), we find a point ($l_{v_i}^\mathsf{r}$) on the road border which has the minimum Euclidean distance to the vehicle's central location. This distance is shown by the red dash-line in Figure \ref{fig-satsegang} which is perpendicular to the road border.

We consider a small circle (the blue circle) with negligible radius $\mathfrak{r}$ centring at $l_{v_i}^\mathsf{r}$.
Then, we find locations of two points ($\Psi_{1}$ and $\Psi_{2}$ ), in which the circle intersects the road boundary. Finally, similar to a derivative operation, the heading angle is calculated by $\theta(\Psi_{1} , \Psi_{2})$, which %calculates 
represents the slope of the red lines at the road boundary, as well as the vehicle heading angle (Figure \ref{fig-satsegang}).\\

%>>>>>>>>>>>>>>>>>>>>>>>>>>>
%\subsubsection{3D Bounding Boxes} \label{bbox-section}
\subsubsection{2D to 3D Bounding Box Conversion} \label{bbox-section}

\begin{figure}[t!]
\centering
\subfloat[Detected objects in 2D bounding boxes]{
\resizebox{8.7cm}{!}{\includegraphics{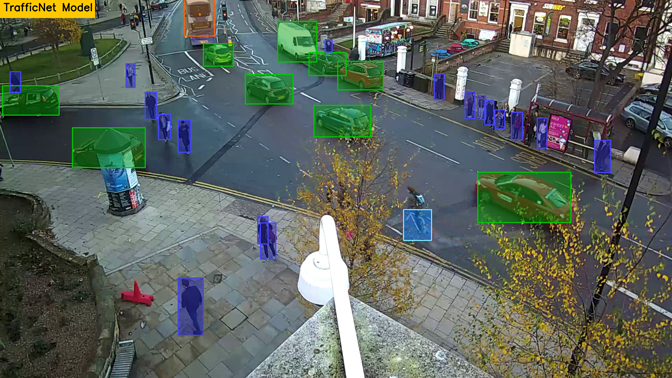}\label{bbox2d}}}\vspace{1pt}
\subfloat[2D to 3D bounding box conversion]{
\resizebox{8.8cm}{!}{\includegraphics{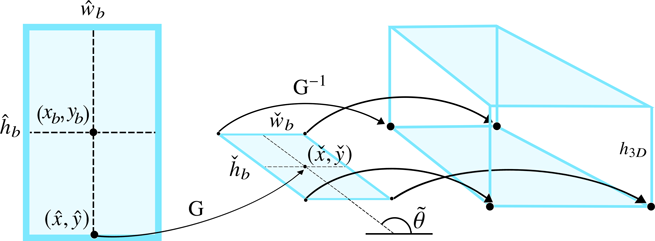}\label{create3d}}}\vspace{1pt}
\subfloat[Final 3D representation]{
\resizebox{8.7cm}{!}{\includegraphics{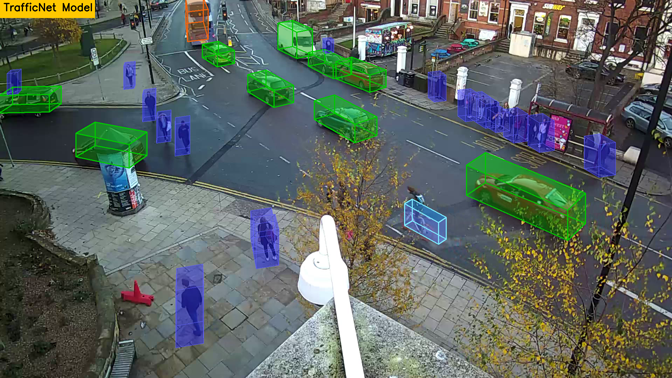}\label{bbox3d}}}\vspace{1pt}
\caption{2D to 3D bounding box conversion process in four categories of vehicle/truck, pedestrian, bus, and cyclist. %  The process of calculating the 3D bounding box based on detected 2D bounding box. Dark blue boxes are pedestrians, light blue is cyclist, greens are cars and orange is bus.
}\label{fig-3D}
\end{figure}

%[fm] modified:  
In order to determine the occupied space of each object in $\hat{\textbf{I}}$ domain, we convert a 2D bounding box (Figure~\ref{bbox2d}) to a cubical 3D bounding box by estimating 8 cube's corners. 
The cube's floor consists of $4$ corner points and corresponds to a rectangle in the $\check{\textbf{I}}$ domain (Figure~\ref{create3d} the middle shape). 
This rectangle indicates the area of the ground plate which is occupied by the object, and can be addressed with the centre $(\check{x},\check{y})$, the height $\check{h}_b$ and the width $\check{w}_b$.
The $\check{h}_b$ and $\check{w}_b$ are determined based on prior knowledge about the approximate height and width of the corresponding object's category in real world (i.e. $5.80 \times 2.9$ meter for buses in the UK). In order to have these distances in pixel criterion, we divide them by the pixel-to-meter ratio ($\iota$), as explained in the Speed Estimation section (\ref{Sec.speed.method}). 

For each vehicle, the rectangle is rotated and aligned with the estimated heading angle $\tilde{\theta}_{v_i}$ to represent the object's movement direction. Then, the four corners of the resulting rectangle are converted to  $\hat{\textbf{I}}$ domain using the $\textbf{G}^{-1}$ matrix and considered as the corners of the cube's floor.
Afterwards, we add $h_{3D}$ to the $y$ axis of the floor corners to indicate the $4$ points of the cube's roof.

The height of the cube
% added
for all road users, except the pedestrians, is set $h_{3D} = \beta \times {h}_b$, % for vehicles
where $\beta = 0.6$ is determined by our experiments as a suitable height coefficient for the detected bounding boxes% height 
in the $\hat{\textbf{I}}$ domain.
%[fm] added:
The cube's height for pedestrians is equal to the height of the detected bounding box in the perspective domain ($h_{3D} = {h}_b$).

Figure~\ref{bbox2d},~\ref{create3d},~\ref{bbox3d} show the hierarchical steps of our approach from 2D to 3D conversion on Leeds University Laidlaw Library surveillance camera footage. %We can see the 3D bounding box estimation process in Figure~\ref{create3d}, and the final results in Figure~\ref{bbox3d}.

%=========================================
\section{Experiments}\label{experiments}

In this section, we evaluate the performance and accuracy of the proposed 3D road-users detection model followed by assessing the efficiency of the proposed environment modelling. 

\subsection{Performance Evaluation} \label{obj-section}

%%%%%%%% DATA
The majority of modern object detectors are trained and evaluated on large and common datasets such as Microsoft Common Objects in Context (Ms-COCO) \cite{lin2015microsoft}. The COCO dataset consists of $886,284$ samples of general annotated objects for $80$ categories (include person, animal, appliance, vehicle, accessory etc.) in $123,287$  images. However, none of them is dedicated to traffic monitoring purposes. 

We considered the MIO-TCD dataset \cite{Luo2018} which consists of $648,959$ images and $11$ traffic-related annotated categories (including cars, pedestrian, bicycle, bus, three types of trucks, two types of vans, motorised vehicles, and non-motorised vehicles) to train and evaluate our models. The dataset has been collected at different times of the day and different %periods 
seasons of the year by thousands of traffic cameras deployed all over Canada and the United States.

As per Table \ref{testset}, we also considered two more traffic monitoring video-footage including UA-DETRAC \cite{UA-DETRAC2020} and GRAM Road-Traffic Monitoring (GRAM-RTM)\cite{guerrero2013iwinac} to test our models under different weather and day/night lighting conditions. 
Moreover, we set up our own surveillance camera at one of the highly interactive intersections of Leeds City, near the Parkinson Building at the University of Leeds, to further evaluate the performance of our model on a real-world scenario consisting of $940,000$ video frames from the live traffic.

\begin{table*}[t!]
\renewcommand\arraystretch{1.2}
\footnotesize % You need these 2 lines to fit the table in page width
\centering
\caption{Specifications of the test datasets used in this research, including various weather conditions, resolutions, frame rates, and video lengths.}
\begin{tabular}{ c c c c c } 
\hline
\multirow{2}{*}{Dataset}  & \multirow{2}{*}{Weather} & \multirow{2}{*}{Length (frame)} & \multirow{2}{*}{Resolution}& \multirow{2}{*}{\textit{fps}}  \\
  & & & & \\ \hline 
\rule{0pt}{3ex} UA-DET \cite{UA-DETRAC2020}   & Sunny, Rainy, Cloudy, Night & 140000 & $960 \times 540$ & 25\\
GRAM-RTM \cite{guerrero2013iwinac}  & Sunny, Foggy  & 40345 & $1200 \times 720$ & 30\\
UK Leeds Parkinson & Day, Sunset, Night & 940000 & $1920 \times 1080$ & 30 \\ \hline
\end{tabular}
\label{testset}
\end{table*}

\begin{figure*}[t!]
\centering
\subfloat[Train]{
\resizebox{17cm}{!}{\includegraphics{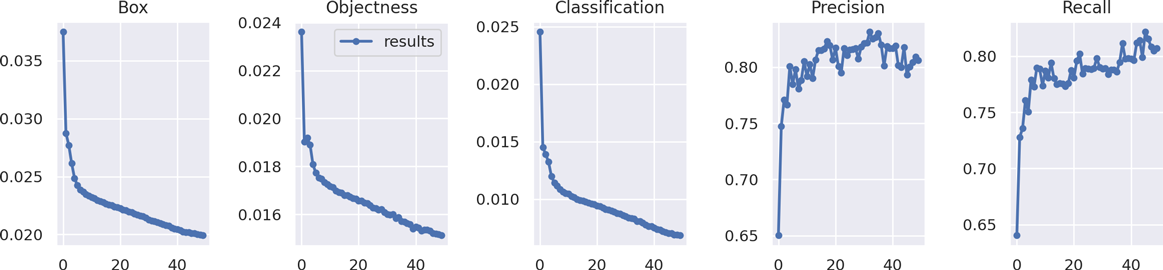}\label{train}}}\vspace{1pt}
\subfloat[Validation results]{
\resizebox{17cm}{!}{\includegraphics{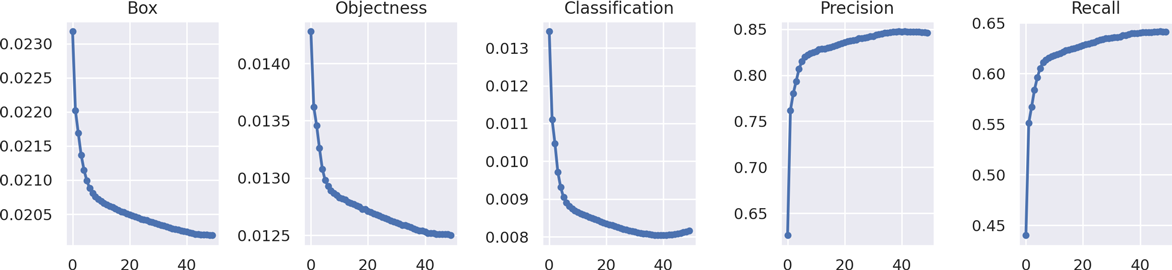}\label{valid}}}
\caption{ Error minimisation graphs of the model in training and validations phases, after 50 epochs.} \label{train-valid}
\vspace{-1mm}
\end{figure*}

%%%%%%%% MODEL
As mentioned in the Methodology section (\ref{det-section}), we adopted transfer learning to train different architectures of YOLOv5 model on the MIO-TCD dataset. We exploited pre-trained weights of $80$ class COCO dataset as initial weights of our fine-tuning process. 

There are four versions of YOLOv5 which are distinguished by the number of learning parameters. The ``small'' with $7.5$ million parameters is a lightweight version, ``medium'' version ($21.8$ million), ``large'' ($47.8$ million), and ``xlarg'' version which has $89$ million learnable parameters. We performed experiments with different number of head modules which consist of three or four head outputs to classify different sizes of objects (as described in section \ref{det-section}).

In the training phase (Figure \ref{train}), we minimised the loss function of the adapted YOLOv5, based on a sum of three loss terms including the "C-IoU loss" as the bounding box regression loss, "objectness confidence loss", and "binary cross entropy" as the classification loss. 

In order to choose optimal learning-rate and avoid long training time, we used one-cycle-learning-rate \cite{smith2018disciplined}.
This gradually increases the learning rate to a certain value (called warm-up phase) followed by a decreasing trend to find the minimum loss, while avoiding local minima. In our experiments, we found the minimum and maximum learning rates of $0.01$ and $0.2$ as the optimum values.

Figure \ref{train-valid} illustrates the analytic graphs of the training and validation processes.
As per the classification graphs ( Fig. \ref{train}), the training loss starts decreasing around epoch $35$, while the validation loss starts increasing (Fig. \ref{valid}). This is a sign of over-fitting in which the model starts memorising the dataset instead of learning generalised features. To avoid the effects of over-fitting, we choose the optimal weights which yield the minimum validation loss.
 
Table~\ref{comparison} compares the performance of the proposed YOLOv5-based model with 10 other state-of-the-art object detection method on the challenging dataset of MIO-TCD. Two metrics of \textit{mAP} and speed (\textit{fps}) are investigated. 

As can be seen, the adapted YOLOv5-based model has achieved a considerable increase in mean average precision comparing to the former standard YOLOv4 algorithm ($84.6\%$ versus $80.4\%$ ). The experiments also proved that $3$-head versions of YOLOv5 provides more efficiency in traffic monitoring than the $4$-head versions. The lightweight version of YOLOv5 reaches the highest rate of speed ($123 fps$). 
While the model has sacrificed the accuracy by $-1.7\%$ in comparison to the highest rate ($84.6\%$).

The YOLOv5 xLarge and Large, with 3 heads reach the highest accuracy of $84.6\%$ on the MIO-TCD benchmark dataset.
Although the xLarge model has more parameters to learn features, the network complexity is greater than what is required to learn the features in the dataset. This prevents the accuracy to go beyond $84.6\%$. Also, it suffers from the lack of adequate speed to perform in real-time performance. Whereas the 3-head YOLOv5 Large, has the same $mAP$ score, and provides a real-time performance of $36.5 fps$. This makes the model more suitable for the cases in which heavy post-processing procedures are involved.

\begin{table*}[t!]
\renewcommand\arraystretch{1.2}
\footnotesize 
\centering
\caption{A comparison of mean average precision (mAP) rate between the developed models and 10 other models on MIO-TCD dataset. The accuracy scores of $3$ truck categories (Articulate Truck, Pickup Truck and Single Unit Truck) is averaged and presented in a one column- "Trucks $\times$ 3". }
\renewcommand\arraystretch{1.2}
\resizebox{\linewidth}{!}{%
\begin{tabular}{c  c  c | c c c c c c c c c} 
%\toprule
\multirow{2}{*}{\textbf{Method}}  & 
\multirow{2}{*}{\parbox{1.4 cm}{\textbf{Speed (fps)}}} & 
\multirow{2}{*}{\textbf{mAP}} & 
\multirow{2}{*}{Bicycle} & 
\multirow{2}{*}{Bus}  &
\multirow{2}{*}{Car}  &
\multirow{2}{*}{Motorcycle}   &
\multirow{2}{*}{\parbox{1.2cm}{Motorised\\ Vehicle}} &
\multirow{2}{*}{\parbox{1.8cm}{Non-motorised\\ Vehicle}} &
\multirow{2}{*}{Pedestrian}  &
\multirow{2}{*}{\parbox{1.2cm}{Work Van}}&
\multirow{2}{*}{\parbox{1.3cm}{Trucks $\times$ 3}}\\
&&&&&&&&&&
\\
\hline 
Faster-RCNN \cite{Luo2018} & 9  &70.0	\% &78.3\% &95.2\% &82.6\%  &81.1\% &52.8\%&37.4\% &31.3\% &73.6\%& 79.2\%\\
RFCN-ResNet-Ensemble4 \cite{Jung2017} &- & 79.2\% &  87.3\% & 97.5\% &89.7\% &88.2\% &\textbf{62.3\%} &59.1 \% &48.6 \%& 79.9  \%& 86.4\%\\
SSD-512 \cite{Luo2018} &16 & 77.3\% &  78.6\% & 96.8\% &94.0\%  & 82.3\% & 56.8\%& 58.8\% & 	43.6\% &  80.4\%& 86.4 \%\\
Context ModelA \cite{Wang2017} &- & 77.2\% & 79.9\% & 96.8\% &93.8\%  &83.6\% & 56.4\%& 58.2\% & 42.6\% & 79.6\%& 86.1\%\\
Adaptive Ensemble \cite{hedeya2020super} &- & 74.2\% & 82.2\% & 95.7\% &91.8\%  &87.3\% & 60.7\%& 45.7	\% & 47.9\% &63.8\%& 80.5\%\\
SSD-300 \cite{Luo2018} &16 & 74.0\% & 78.3\% &95.7\% &91.5\% &78.9\% &51.4\%&55.2\% & 37.3\% &75.0\%& 83.5\%\\
YOLOv2-MIOTCD \cite{Luo2018} &18 &71.8\% &78.6\% &95.1\% &81.4\%  &81.4\% &51.7\%&56.6\% &25.0\% & 76.4\%& 81.3\%\\
YOLOv2-PascalVOC \cite{Luo2018} &18 &71.5\% &78.4\% &95.2\% &80.5\%  &80.9\% &52.0\%&56.5\%& 25.7 \%& 75.7\%& 80.4 \%\\
YOLOv1 \cite{Luo2018}&19 &62.7	\% &70.0\% &91.6\% &77.2\%  &71.4\% &44.4\%&20.7\% &18.1\% &69.3\%& 75.5\%\\
 \hline
 \multicolumn{12}{c}{\textbf{Our Experiments}} \\
 \hline
YOLOv4 & 24 & 80.4\%  &  89.2\%  & 95.8\% &91.6\%  & 91.5\% & 58.6\%& 63.9\% & 63.4\% &  79.0\%& 83.7 \%\\
YOLOv5 Small (3 head) & \textbf{123.5}  &82.8\%  &91.6\% &98.3\% &95.5\%  &94.1\% &50.5\%&65.6\% &70.1\%  &81.8\%& 87.8 \%\\
YOLOv5 Medium (3 head) &60.60  &84.1\% &92.4\% &98.4\% &95.9\%  &\textbf{94.3\%} &51.7\%&68.8\%& 74.8\% &83.3\%& 88.6  \% \\
YOLOv5 Large (3 head) &\hl{36.50}  & \hl{84.6\%} &92.5\% &\hl{98.7\%} &95.9\%  &\hl{94.3\%} &51.7\%&70.1\% &\hl{77.4\%} &\hl{83.8\%}& \hl{88.8\%}\\
YOLOv5 xLarge (3 head) &20.16 & \hl{84.6\%} &92.7\% &\hl{98.7\%}& \hl{96.0\%}& 94.1\% &51.7\%&\hl{71.2\%} &76.2\% &\hl{83.8\%}& \hl{88.8\%} \\
YOLOv5 Large (4 head) &117.6  &80.9\% &91.2\% &97.8\% &95.1\%  &91.7\% &48.4\%&61.9\% &64.3\% &80.0\%& 86.6\% \\
YOLOv5 Medium (4 head) &54.90  &82.9\% &92.2\% &98.4\% &95.5\%  &93.1\% &50.0\%&66.8\% &69.5\% &82.1\%&88.1\% \\
YOLOv5 Large (4 head) &33.00 &83.4\% & \textbf{92.9\%} &98.4\% &95.7\%  &93.7\% &50.6\%&68.0\% &71.3\% &82.6\%&88.0\% \\
YOLOv5 xLarge (4 head) &19.20 &83.7\% &91.8\% &98.4\% &95.7\%  &93.5\% &50.8\%&69.0\% &72.2\% &83.4\%&88.5\% \\
\bottomrule
\end{tabular}
}
\vspace{3mm}
\label{comparison}
\end{table*}

% [+] Added:
\begin{table}[t!]
\renewcommand\arraystretch{1.2}
\footnotesize
\centering
\caption{Detection performance of our YOLOv5 Large (3 head) model on two auxiliary traffic-related datasets.}
\begin{tabular}{ c c c  } 
\hline
\multirow{2}{*}{Datasets}  & \multirow{2}{*}{Precision} & \multirow{2}{*}{Recall}   \\
  & & \\ \hline 
\rule{0pt}{3ex} UA-DET \cite{UA-DETRAC2020}   & 99.8\% & 99.7\% \\
GRAM-RTM \cite{guerrero2013iwinac}  & 99.7\%  & 99.5\% \\
\end{tabular}
\label{test2}
\end{table} 

Table \ref{test2} shows the test results of our pioneer detection model (YOLOv5-Large, 3 head) on UA-DET and GRAM-RTM datasets with very high precision rates of 99.8\% and 99.7\%, respectively. The GRAM-RTM dataset only provides ground truth annotations for one lane of the road. So, we applied a mask to ignore the non-annotated lanes of the road; otherwise, our proposed model is capable of detecting vehicles in both lanes.

\begin{figure*}[t!]
\vspace{-1mm}
\centering
\subfloat[Detected cars and bus on frame 82000]{
\resizebox{8.65cm}{!}{\includegraphics{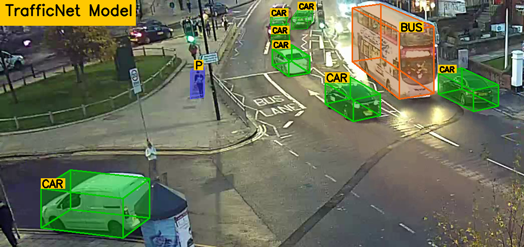}\label{occ3D_1}}}\hspace{1pt}
\subfloat[Detected cars and bus on frame 82100]{
\resizebox{8.65cm}{!}{\includegraphics{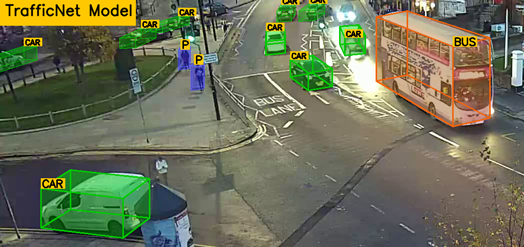}\label{occ3D_2}}}\vspace{1pt}
\subfloat[Environment mapping result of frame 82000]{
\resizebox{8.65cm}{!}{\includegraphics{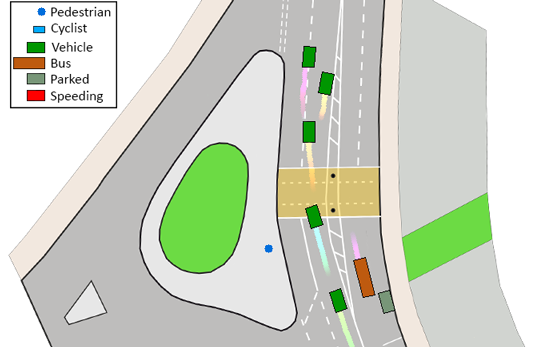}\label{occMap_1}}}\vspace{1pt}
\subfloat[Environment mapping result of frame 82100]{
\resizebox{8.65cm}{!}{\includegraphics{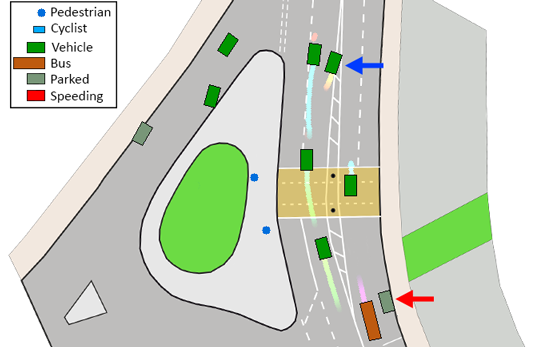}\label{occMap_2}}}\hspace{1pt}
\caption{The outputs of adapted 3-head YOLOv5-large algorithm for road-user detection and environment modelling.}
\label{occlusion}
\end{figure*}

Figure \ref{occlusion}, top row, shows the results of the detection algorithm and 3D localisation of the road users. Figure \ref{occlusion}, the bottom row, shows the environment modelling of the scene as a digital twin of the scene and live traffic information. Such live information (which can be stored in cloud servers), would be be very useful for city councils, police, governmental authorities, traffic policy makers, and even as extra source of processed data for automated vehicles (AVs) which traverse around the same zone. Such rich digital twins of the road condition can significantly along with the ego-vehicles sensory data can enhance the AVs' capability in better dealing with the corner cases and complicated traffic scenarios.

% Added
In Figure \ref{occlusion} we are also trying to show the efficiency of the heading angle estimation and the tracking system in case of full occlusions. As can be seen, one of the cars in the scene is taking a U-turn and we have properly identified the heading angle of the car at frame 82100 (indicated with blur arrow). This can be compared with its previous angle and position in frame 82000. Considering the position and the heading angle of the vehicle at frames 82000 and 82100, the 3D bounding box of the vehicle is also determined. 

	As another complicated example in the same scene, one of the cars is fully occluded by a passing bus at frame 82100 (indicated with a red arrow). However the car has been fully traced by utilisation of the spatio-temporal information and tracking data at frame 82000 and beyond. 

%shows a car indicated with a blue arrow at frame 82100 which is going to make a U-turn. Considering the position of this vehicle at frames 82100 and 82000, the heading angle of the vehicle and its 3D bounding box is determined.

%>>>>>>>>>>>>>>>>>>>>>>>>>>>
\subsection {Environment Modelling and Traffic Analysis}

In order to take the most of the detection and tracking algorithms and to provide smart traffic monitoring analysis, we defined three possible states for vehicles and pedestrians as follows:

\begin{itemize}
\item \textbf{Parking}: a set $\mathcal{P} $ contains all of the vehicles which have less than one-meter distance in $\check{\textbf{I}}$ domain from the road border ($l_{v_i}^\mathsf{r}$), and their 
temporal 
speeds ($\vartheta_{v_i}$) have been close to zero for more than 1 minute.

\item \textbf{Speeding Violation}: a set $\mathcal{S}$ consists of vehicles in which their speed  ($\vartheta_{v_i}$) is more than the speed limit of the road (i.e. 30 \textit{mph} for Leeds city centre, UK). 

\item \textbf{Collision Risk}: a set $\mathcal{D}$ consists of pedestrians whose distances from vehicles are less than a meter, and the vehicles are not in the parking status $\mathcal{P} $.
\end{itemize}

\begin{figure*}[t!]
\centering
\subfloat[Vehicle and pedestrian counts over 6000 video frames. Source: Parkinson building CCTV surveillance camera.]{
\resizebox{17cm}{!}{\includegraphics{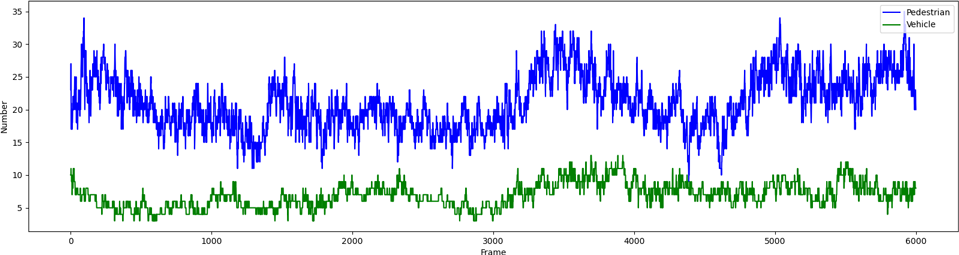}\label{graph-counter}}}\hspace{1pt}
\subfloat[BEV Pedestrian movements heat map]{
\resizebox{8.65cm}{!}{\includegraphics{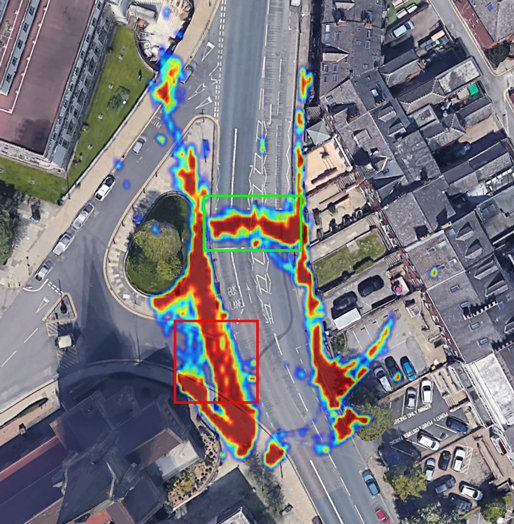}\label{fig-leeds-hpm-bev}}}\vspace{1pt}
\subfloat[BEV vehicle movements heat map]{
\resizebox{8.65cm}{!}{\includegraphics{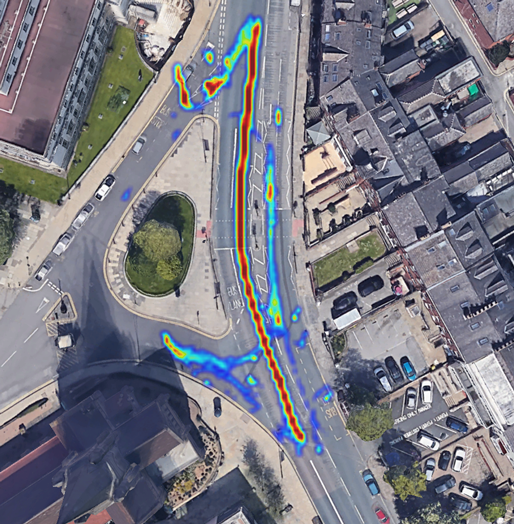}\label{fig-leeds-hvm-bev}}}\hspace{1pt}
\vspace{1pt}
\subfloat[Pedestrian movements heat map- Perspective view]{
\resizebox{8.65cm}{!}{\includegraphics{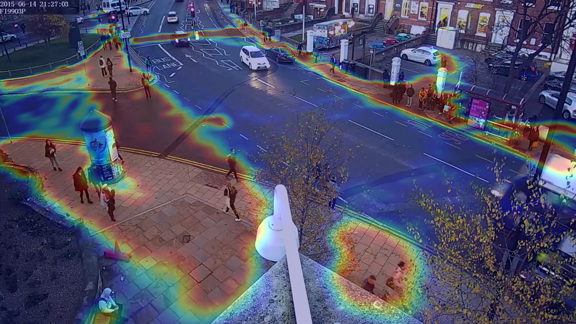}\label{fig-leeds-hpm}}}
\subfloat[Vehicle movements heat map- Perspective view]{
\resizebox{8.65cm}{!}{\includegraphics{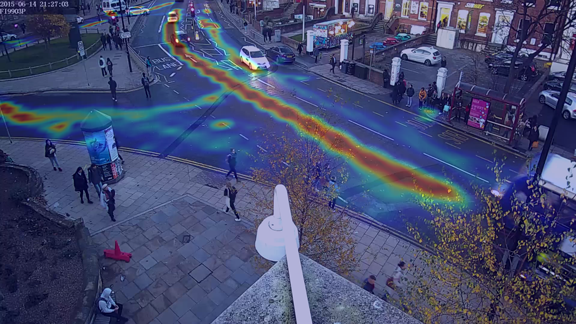}\label{fig-leeds-hvm}}}\hspace{1pt}
\caption{Spatio-temporal long-term analysis of vehicles and pedestrians' activity using Parkinson Building surveillance camera, Leeds, UK}\label{fig-movements}
\end{figure*}

To analyse the traffic condition, we buffer the count of tracked vehicles and pedestrians locations during a period of time (e.g. 6,000 frames) as shown by line graph in Figure \ref{graph-counter}.

In order to visualise a long-term spatio-temporal statistical analysis of traffic flow and interactions between road users, a heat map representation is created similar to our previous work in another context for social distancing monitoring~\cite{rezaei2020deepsocial}.
The heat map is defined by the matrix $\check{\textbf{H}}^t \in \mathbb{R}^{\check{w} \times \check{h}}$ in the satellite domain, where $t$ is the frame-ID number.
The matrix is initially filled with zero to save the last location of objects using the input image sequences.
The heat map updates in each frame by the function $G_{(\text{object})}(\check{\textbf{H}})$ which applies a $3 \times 3$ Gaussian matrix centred at the object's location ($\check{x}, \check{y}$) on the $\check{\textbf{H}}$ matrix. Finally, we normalise the heat map intensity values between 0 and 255, in order to visualise it as a colour-coded heat image. Then a colour spectrum will be mapped to the stored values in the $\check{\textbf{H}}$ matrix in which the red-spectrum represents the higher values, and the blue-spectrum represents the low values.\\

The heat map of the detected pedestrians is shown by $\check{\textbf{H}}_{(p)}$, which updates over time as follows:
\begin{equation}\label{pmov}
\check{\textbf{H}}^t_{(p)} = G_{(p_i)}(\check{\textbf{H}}^{t-1}_{(p)}) \hspace{0.5cm} \forall {p_i} \in P \\
\end{equation}
Figure \ref{fig-leeds-hpm-bev} illustrates the developed heat map $\check{\textbf{H}}_{(p)}$on the satellite image. The lower range values have been removed for better visualisation. 
This figure provides valuable information about the pedestrians' activity. For instance, we can see a significant number of pedestrians have crossed the dedicated zebra-crossing shown by the green rectangle. However, in another region of the road (marked by a red rectangle) many other pedestrians cross another part of the road where there is no zebra-crossing. 
Also, there are a few pedestrians who has crossed the street directly in front of the bus station.

Similarly, the heat map for detected vehicles is defined as follows:
\begin{equation}\label{vmov}
\check{\textbf{H}}^t_{(v)} = G_{(v_i)}(\check{\textbf{H}}^{t-1}_{(v)} ) \hspace{0.5cm} \forall {v_i} \in V \hspace{0.1cm} ,\hspace{0.1cm} v_i \notin \mathcal{P}
\end{equation}
\noindent where $\check{\textbf{H}}_{(v)}$ stores the location of moving vehicles only (not stationary or parked vehicles). 
This matrix has illustrated in Figure\ref{fig-leeds-hvm-bev}.
This heat map represents that more vehicles are traversing on the left lane of the road comparing to the opposite direction, on the right lane.

The heat map images can be also mapped to the perspective space by: $\hat{\textbf{H}} =\Lambda(\check{\textbf{H}} , \textbf{G}^{-1})$. Figures \ref{fig-leeds-hpm} and \ref{fig-leeds-hvm} are corresponded maps of Figures \ref{fig-leeds-hpm-bev} and \ref{fig-leeds-hvm-bev}, respectively.

\begin{figure}[t!]
\centering
\subfloat[Speed violation heat map- Perspective view]{
\resizebox{8.55cm}{!}{\includegraphics{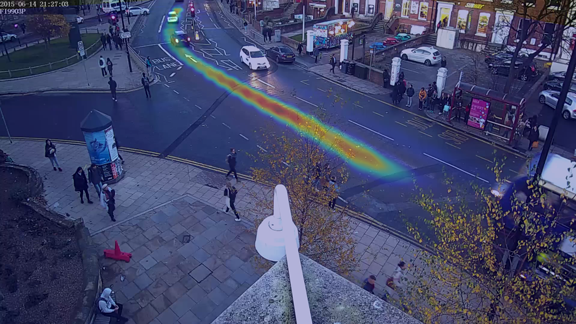}\label{fig-leeds-hsv}}}
\vspace{1pt}
\subfloat[BEV speed violation heat map]{
\resizebox{8.65cm}{!}{\includegraphics{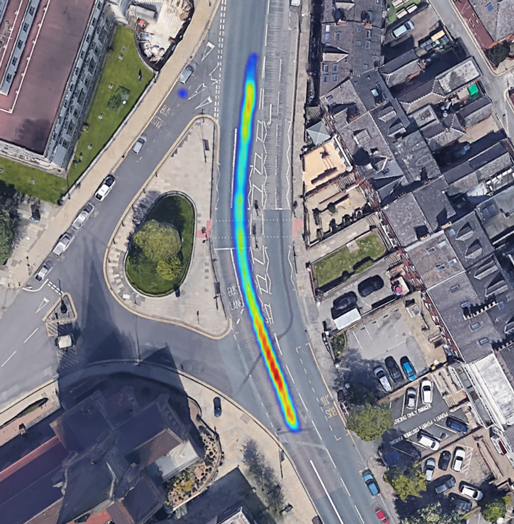}\label{fig-leeds-hsv-bev}}}\hspace{1pt}
\subfloat[Average speed of moving vehicles in the scene]{
\resizebox{8.65cm}{!}{\includegraphics{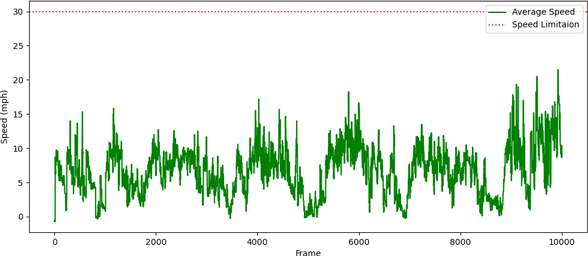}\label{fig-avgs}}}\vspace{1pt}
\vspace{-1mm}
\caption{Automated speed monitoring and heat map analysis based on 10,000 video frames from the Laidlaw Library surveillance camera, Leeds, UK}\label{avgs}
\vspace{-3mm}
\end{figure}

We also investigated the speed violation heat map $\check{\textbf{H}}_{(\vartheta)}$ and the areas in which vehicles violated the speed limit of the road:
\begin{equation}\label{speeding}
\check{\textbf{H}}^t_{(\vartheta)} = G_{(v_i)}(\check{\textbf{H}}^{t-1}_{(\vartheta)}) \hspace{0.5cm} \forall {v_i} \in \mathcal{S} 
\end{equation}
Figure %\ref{avgs} 
\ref{fig-leeds-hsv} and \ref{fig-leeds-hsv-bev}, illustrates an instance of speed heat map calculated over the 10,000 selected frames. 
As can be seen the speeding violation significantly decreases near the pedestrian crossing zone, which makes sense. As a very useful application of our developed model, similar investigations can be conducted in various parts of city and urban areas, in order to identify less known or hidden hazardous zones where the vehicles may breach the traffic rules.

The graph shown in Figure \ref{fig-avgs}, represents the average speed of all vehicles in the scene during the selected period of the monitoring. % is shown in Graph\ref{fig-avgs}.
In each frame,  the average speed is calculated by: 
\begin{equation}
\bar{\vartheta} =\frac{\sum \vartheta_{v_i}}{n_v}  \hspace{0.4 cm} \forall  v_i \in V, v_i \notin \mathcal{P}
\end{equation}
\noindent where $n_v$ is the number of vehicles that are not in the Parking state. 
 
In order to identify the congested and crowded spots in the scene, we can monitor the vehicles e.g. with less than 2\textit{m} distances to each other with an average speed of e.g. lower than 5\textit{mph}. The shorter vehicles' proximity over a longer period of time, the larger values will be stored in the congestion buffer; consequently, a hotter heat map will be generated. Defining optimum values of distance and speed threshold requires and intensive analytical and statistical data collection and assessments based on the road type (e.g. highway or a city road) which is out of the scope of this research.  

\begin{figure}[t!]
\centering
\includegraphics[width = 1\linewidth]{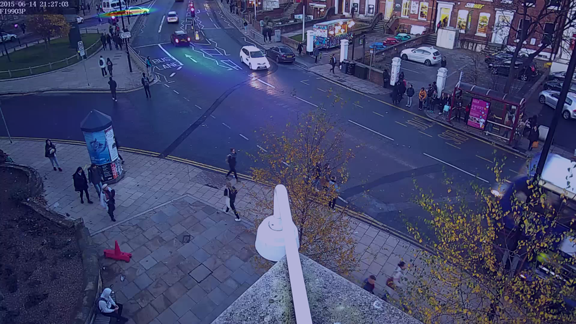}
\vspace{-5mm}
\caption{Heat map representation of congested areas based on 10,000 live video frames from Woodhouse lane, Leeds LS2 9JT, UK.}
\vspace{3mm}
\label{fig-leeds-crowd}
\end{figure}

However, as a general-purpose solution and similar to the previous heat maps, we defined the congestion heat map $\check{\textbf{H}}_{(\mathcal{C})}$ as follows:
\begin{equation}\label{congested}
\check{\textbf{H}}^t_{(\mathcal{C})} = G_{(v_i)}(\check{\textbf{H}}^{t-1}_{(\mathcal{C})}) \hspace{0.5cm} \forall {v_i} \in \mathcal{A} 
\end{equation}

\noindent where $\mathcal{A}$ is an ID set of vehicles that are in the congested areas.
As we can see in Figure~\ref{fig-leeds-crowd}, there are two regions of congestion, one before the pedestrian crossing which is probably due to the red traffic light which stops the vehicles, and also a second congestion spot at the T-junction (top left side of the scene), where the vehicles stop and line up before joining the main road.

\begin{figure}[t!]
\centering
\includegraphics[width = 1\linewidth]{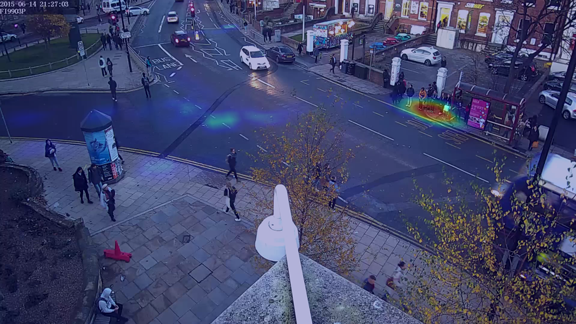}
\vspace{-5mm}
\caption{Heat map representation of areas in which vehicles and pedestrians were too close to each other. Source/Location: 10,000 live video frames, Woodhouse lane, Leeds LS2 9JT, UK.}
%\vspace{3mm}
\label{fig-leeds-hnp}
\end{figure}

%___________danger assessment
Figure~\ref{fig-leeds-hnp} shows the pedestrian behaviour's heat map %of the 
by monitoring the
pedestrians who are not maintaining a minimum safety distance of $2m$ to the passing vehicles. % (i.e. pedestrian behaviours).
Similarly, the heat map of the high-risk pedestrians can be updated according to the following equation:
\begin{equation}\label{too-close}
\check{\textbf{H}}^t_{(\mathcal{W})} = G_{(p_i)}(\check{\textbf{H}}^{t-1}_{(\mathcal{W})}) \hspace{0.5cm} \forall {p_i} \in \mathcal{D} 
\end{equation}

The hot area in front of the bus station is more likely caused by the buses which stop just beside the bus station. The heat map also shows another very unsafe and risky spot in the same scene where some of the pedestrians have crossed through the middle of a complex 3-way intersection. This may have been caused by careless pedestrians who try to reach the bus stop or leave the bus stop via a high-risk shortcut.  

All experiments and performance evaluations in this research were conducted on a PC workstation with an Intel \copyright Core\texttrademark{} i5-9400F processor and an NVIDIA RTX 2080 GPU with CUDA version~11. All services were performed based on a unified software using parallel processing for simultaneous utilisation of all processor's cores to enhance the execution performance. Similarly, all image-processing-related calculations were performed on GPU tensor units to increase speed and efficiency. 

The %performance 
running time of the whole services is 0.05 $ms$, except for the speed of the object detector which can slightly vary depending on the lighting and complexity of the environment.

%=========================================
\section{Conclusion}\label{conc}

In this article, we proposed a real-time traffic monitoring system called Traffic-Net which applies a customised 3-head YOLOv5 model to detect various categories of vehicles and pedestrian.
A multi-class and multi-object tracker named MOMCT were also developed for an accurate and continuous classification, identifications, and localisation of the same objects over consequent video frames, as well as prediction of the next position of vehicles in case of missing information.
In order to develop a general-purpose solution applicable on the majority of traffic surveillance cameras, 
we introduced an automatic camera calibration techniques (called SG-IPM) to estimate real-world positions and distances using a combination of near-perpendicular satellite images and ground information.

Having the real-world position 
% added
of the vehicles, 
a constant acceleration Kalman filter was applied for smooth speed estimation. Using spatio-temporal moving trajectory information, the heading angle of vehicles were also calculated. We also introduced the ABF method to remove the angle variation noise due to occlusion, sensor limitation, or detection imperfection. 

These led to 3D bounding box estimation and traffic heat map modelling and analysis which can help the researchers and authorities to automatically analyse the road congestion, high-risk areas, and the pedestrian-vehicle interactions. Experimental results on the MIO-TCD dataset and a real-world road-side camera, confirmed the proposed approach well dominates 10 state-of-the-art research work in ten categories of vehicles and pedestrian detection. Tracking, auto-calibration, and automated congestion detection with a high level of accuracy (up to 84.6\%) and stability over various lighting conditions were other outcomes of this research. 

As a future study and in order to improve the feature matching process between the camera and satellite images, a neural network-based feature matching algorithm can be applied to increase the accuracy. Also, many other strategies (like evolutionary algorithms, feature engineering, and generative models) can be used to provide more robust features, to tackle the matching failures. 

Availability of larger datasets can further help to improve the accuracy of heat maps, to identify high-risk road spots and further statistical analyses.

\section*{Acknowledgement}
The research has received funding from the European Commission Horizon 2020 program under the L3Pilot project, grant No. 723051 as well as the interACT project from the European Union’s Horizon 2020 research and innovation program,  grant agreement No. 723395. Responsibility for the information and views set out in this publication lies entirely with the authors. 

\bibliography{ref}

\newpage
\clearpage
\section*{Appendix 1: Camera Calibration and Inverse Perspective Mapping}

Knowing the camera intrinsic and extrinsic parameters, the actual position of the 3D objects from 2D perspective image can be estimated using Inverse Perspective Mapping (IPM) as follows:

\begin{equation}\label{ipm}
[x\hspace{0.2cm} y \hspace{0.2cm} 1]^T = \textbf{K}[\textbf{R}|\textbf{T}] [X_w \hspace{0.2cm} Y_w \hspace{0.2cm} Z_w  \hspace{0.2cm}1]^T
\end{equation}
\noindent where $x$ and $y$ are the pixel coordinates of the image, $X_w$, $Y_w$ and $Z_w$ are coordinates of points in real world. $\textbf{K}$ is the camera intrinsic matrix: 
\begin{equation}\label{kmat}
\textbf{K}=
\begin{bmatrix}
f * k_x & s & \mbox{c}_x & 0\\
0 & f * k_y & \mbox{c}_y & 0\\
0 & 0 & 1 & 0
\end{bmatrix}
\end{equation}
\noindent where $f$ is the focal length of the camera, $k_x$ and $k_y$ are the calibration coefficient values in horizontal and vertical pixel axis, $s$ is the shear coefficient and $(\mbox{c}_x, \mbox{c}_y)$ are the principal points shifting the optical axis of the image plane.

\noindent $\textbf{R}$ is the rotation matrix:
\begin{equation}\label{rotation}
\textbf{R}=
\begin{bmatrix}
1 & 0 & 0 & 0\\
0 & \cos\theta_{c}& -\sin\theta_{c} & 0\\
0 & \sin\theta_{c} & \cos\theta_{c} & 0\\
0 & 0 & 0 & 1
\end{bmatrix}
\end{equation}
\noindent where $\theta_c$ is the camera angle.

\noindent $\textbf{T}$ is the translation matrix:
\begin{equation}
\label{translation}
\textbf{T}=
\begin{bmatrix}
1 & 0 & 0 & 0\\
0 & 1 & 0 & 0\\
0 & 0 & 1 & -\frac{h_{c}}{\sin\theta_{c}}\\
0 & 0 & 0 & 1
\end{bmatrix}
\end{equation}
\noindent where $h_{c}$ is the height of the camera.

These three matrices together $\textbf{K}[\textbf{R}|\textbf{T}]$ are known as projection matrix $\textbf{G} \in \mathbb{R}^{3 \times 4}$, so the transformation equation can be summarised as $[x \hspace{0.2cm} y \hspace{0.2cm} 1]^T = \textbf{G} \hspace{0.1cm}[ X_w \hspace{0.2cm} Y_w \hspace{0.2cm} Z_w \hspace{0.2cm} 1] ^T$. 

Assuming the camera is looking perpendicular to the ground plane of the scene, the $Z_w$ parameter is removed. A reduction in the $\textbf{G}$ matrix size, turns it into a planar transformation matrix $\textbf{G} \in \mathbb{R}^{3 \times 3}$ with  $g_{ij}$ elements as follows:
\begin{equation}\label{3by3t}
\begin{bmatrix}
x\\
y\\
1
\end{bmatrix}
=
\begin{bmatrix}
g_{11} & g_{12} & g_{13} \\
g_{21} & g_{22} & g_{23} \\
g_{31} & g_{32} & g_{33} \\
\end{bmatrix}
\begin{bmatrix}
X_w\\
Y_w\\
1
\end{bmatrix}
\end{equation}

Therefore, for every pixel point $(x,y)$, the planar transformation function can be represented as follow: 
\begin{equation}\label{func}
\begin{aligned}
	\Lambda((x,y),\textbf{G})=
(\frac{g_{11}\times x+g_{12}\times y+g_{13}}{g_{31}\times x+g_{32}\times y+g_{33}} , \\
\frac{g_{21}\times x+g_{22}\times y+g_{23}}{g_{31}\times x+g_{32}\times y+g_{33}})
\end{aligned}
\end{equation}

\end{document}